\documentclass[10pt,twocolumn,letterpaper]{article}

\usepackage[pagenumbers]{cvpr} % To force page numbers, e.g. for an arXiv version

\usepackage{graphicx}
\usepackage{amsmath}
\usepackage{amssymb}
\usepackage{booktabs}

\usepackage{bbm}
\usepackage{multirow}

\makeatletter
\def\hlinew#1{%
	\noalign{\ifnum0=`}\fi\hrule \@height #1 \futurelet
	\reserved@a\@xhline}
\makeatother%

\usepackage{arydshln}

\usepackage{pifont}

\usepackage{colortbl}
\usepackage{xcolor}

\usepackage{paralist}
\usepackage[pagebackref=true,breaklinks=true,letterpaper=true,colorlinks,citecolor=blue,linkcolor=red,bookmarks=false]{hyperref}

\usepackage[capitalize]{cleveref}
\crefname{section}{Sec.}{Secs.}
\Crefname{section}{Section}{Sections}
\Crefname{table}{Table}{Tables}
\crefname{table}{Tab.}{Tabs.}

\def\cvprPaperID{1098} % *** Enter the CVPR Paper ID here
\def\confName{CVPR}
\def\confYear{2023}

\begin{document}

%%%%%%%%% TITLE - PLEASE UPDATE
\title{DLGSANet: Lightweight Dynamic Local and Global Self-Attention Networks\\ for Image Super-Resolution}

\author{Xiang Li, Jinshan Pan, Jinhui Tang, and Jiangxin Dong\\
Nanjing University of Science and Technology\\
\\
}
% \author{First Author\\
% Institution1\\
% Institution1 address\\
% {\tt\small firstauthor@i1.org}
% % For a paper whose authors are all at the same institution,
% % omit the following lines up until the closing ``}''.
% % Additional authors and addresses can be added with ``\and'',
% % just like the second author.
% % To save space, use either the email address or home page, not both
% \and
% Second Author\\
% Institution2\\
% First line of institution2 address\\
% {\tt\small secondauthor@i2.org}
% }
\maketitle

%%%%%%%%% ABSTRACT
\begin{abstract}
   %
%   In this paper, we propose an effective lightweight residual mix dynamic-Transformer network (DLGSANet) to solve image super-resolution efficiently.
   We propose an effective lightweight dynamic local and global self-attention network (DLGSANet) to solve image super-resolution.
   Our method explores the properties of Transformers while having low computational costs.
   Motivated by the network designs of Transformers, we develop a simple yet effective multi-head dynamic local self-attention (MHDLSA) module to extract local features efficiently.
   In addition, we note that existing Transformers usually explore all similarities of the tokens between the queries and keys for the feature aggregation.
   However, not all the tokens from the queries are relevant to those in keys, using all the similarities does not effectively facilitate the high-resolution image reconstruction.
   To overcome this problem, we develop a sparse global self-attention (SparseGSA) module to select the most useful similarity values so that the most useful global features can be better utilized for the high-resolution image reconstruction.
   We develop a hybrid dynamic-Transformer block (HDTB) that integrates the MHDLSA and SparseGSA for both local and global feature exploration.
   To ease the network training, we formulate the HDTBs into a residual hybrid dynamic-Transformer group (RHDTG).
   %
%   By embedding the RHDTG into an end-to-end trainable network, we show that the proposed method has fewer network parameters and lower computational costs while achieving competitive performance against state-of-the-art ones in terms of accuracy.
    %
    By embedding the RHDTGs into an end-to-end trainable network, we show that our proposed method has fewer network parameters and lower computational costs while achieving competitive performance against state-of-the-art ones in terms of accuracy.
    More information is available at \url{https://neonleexiang.github.io/DLGSANet/}.
\end{abstract}

%%%%%%%%% BODY TEXT
% =====================================================================================
\section{Introduction}
%---
%A sizable amount of computer vision domains are devoted to low-level vision tasks.
%
%In the area of low-level vision task,

Single image super-resolution (SISR) aims to find a solution to the issue of reconstructing a high-resolution image from a low-resolution one so that the high-resolution image can be better displayed on high-definition devices.
In order to produce high-resolution images, classical approaches, e.g., bicubic and bilinear, employ interpolation processes to complement the surrounding pixel values.
Convolutional neural network (CNN)-based approaches such as \cite{SRCNN, FSRCNN, VDSR, EDSR, RCAN} tackle the image super-resolution challenge, generating better super-resolved images than those of conventional approaches.
% as a consequence of the advancement of deep learning brought about by \cite{ResNet}.
%
%These CNN-based techniques have improved scene restoration for some particular scenarios, which can improve the visual quality of super resolution output images.
These CNN-based approaches have greatly advanced the progress of SISR.
%---

% +++++++++++++++++++++++++++++++++++++++++++++++++++++++++++++++++++++++++++++++++++++
\begin{figure}[!t]
	\centering
	\includegraphics[width=0.48\textwidth]{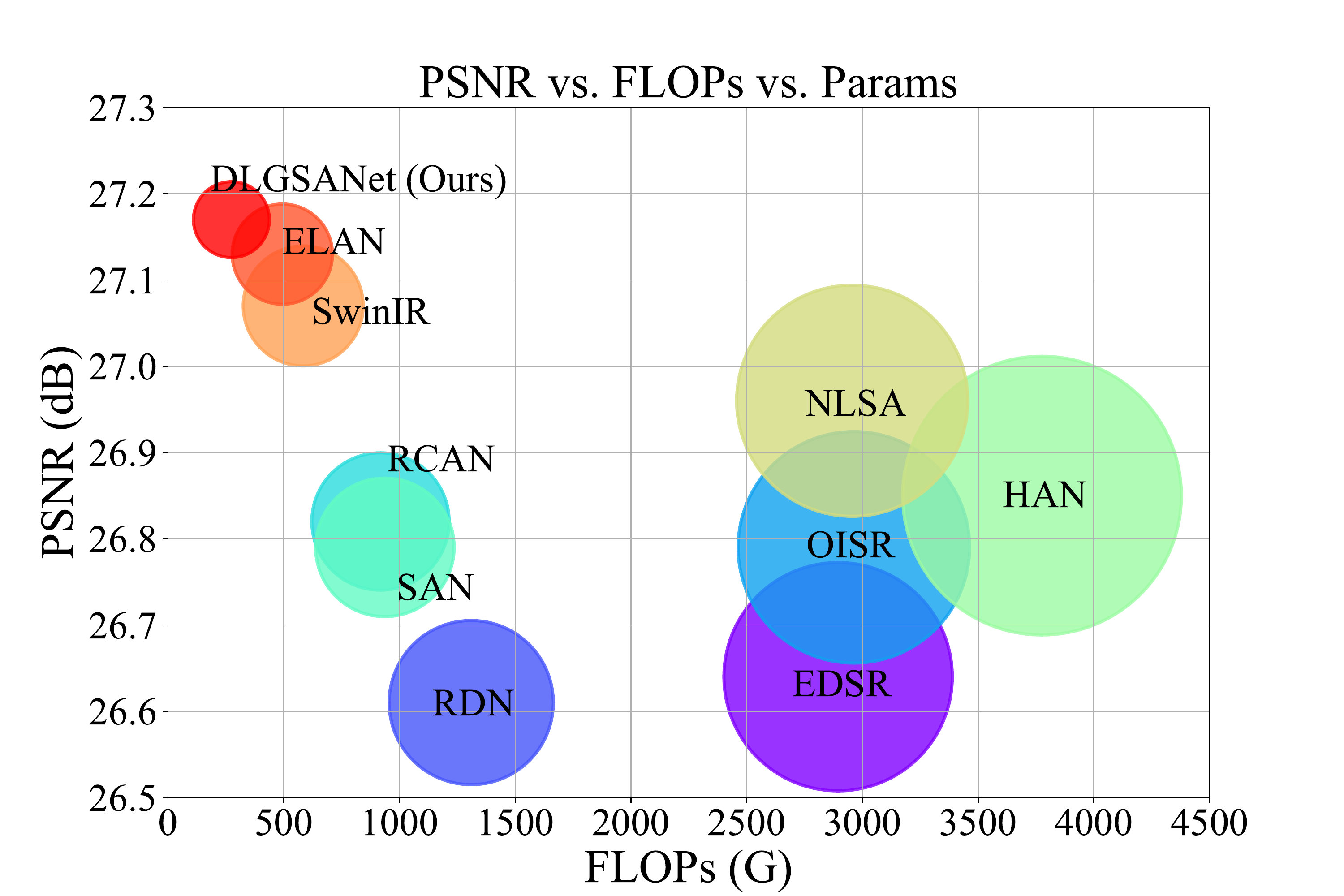} % Reduce the figure size so that it is slightly narrower than the column.
	\vspace{-4mm}
	\caption{Image super-resolution comparisons ($\times 4$) in terms of accuracy, network parameters, and floating point operations (FLOPs) from the Urban100 dataset.
The area of each circle denotes the number of network parameters. Our model (DLGSANet) achieves comparable performance while having fewer network parameters ($<5$M) and lower FLOPs.}
	\label{complexity scatter}
    \vspace{-3mm}
\end{figure}
% +++++++++++++++++++++++++++++++++++++++++++++++++++++++++++++++++++++++++++++++++++++

%---
Furthermore, several follow-up studies, such as \cite{NLSA, HAN, OISR}, progressively start to develop larger and deeper CNN models for better learning capacity.
Although the quality of the super-resolved images is largely improved, the computational costs of those approaches is quite expensive due to the large number of network parameters and calculations (e.g., more than 60M in network parameters and 3000G in FLOPs), which limits their real-world applications.
Thus, there is a great need to develop a lightweight and efficient model to solve SISR.
%%====

%%
%Some related studies concentrate on improving feature extraction.
%%
%For example, \cite{RCAN} can flexibly extract various feature information by re-weighting each channel using a channel attention mechanism.
%%
%Additionally, self-attention (SA) is proposed by \cite{ViTs} for larger receptive fields and more flexibility.
%%
%By converting the global receptive field into a small window, \cite{SwinT} minimizes computing costs while also improving performance.
%%
%In super resolution task, \cite{SwinIR} employs the Swin-Transformer as a building block and achieves superior performance when compared to CNN-based techniques. In addition, \cite{SwinIR} only costs 11M in parameters when compared to \cite{IPT} and other CNN-based methods.
%%
%Thus, self attention can be a good baseline for low-level tasks and can extract stronger features than the majority of the previously proposed CNN-based approaches.
%%---

%%
%Recently, Transformers (ViTs) have shown great success in lots of vision problems and are applied to SISR~\cite{}.
As Vision Transformers (ViTs)~\cite{ViTs} can model global contexts while having fewer network parameters, a recent method~\cite{IPT} applies them to SISR and achieves better results in terms of accuracy and network parameters compared to the CNN-based ones.
However, as the original ViTs are computationally expensive, the shifted window scheme has been adopted in~\cite{SwinT}.
%%
%However, as the original ViTs are computational expensive, the shifted windowing scheme has been developed to alleviate this problem. One of the representative methods, Swin Transformer, has been applied to SISR and achieves decent performance compared to the CNN-based methods.
Although the self-attention by the shifted window scheme is capable of extracting local features, discontinuous windows limit the ability to model local features within each window.
%
% Furthermore, as the window is fixed for various images, these methods must execute a padding operation on the input to make sure that it can be divided into fixed-size windows during the inference.
%
Moreover, the window-based methods are unable to aggregate information outside of the window, which leads to limited ability for modeling global information.

To better explore global features while reducing the computational costs, several approaches, e.g., \cite{Restormer}, develop transposed attentions that compute the self-attention along the number of features. We note that these transformer-based methods usually use all the similarity values in the self-attention for feature aggregation. However, as not all the tokens from the queries are relevant to those in keys, using all similarities does not effectively facilitate the high-resolution image reconstruction.
%
%Therefore,  employs Transposed Attention, a low-cost computing method, to aggregate global information through channel-wise.
Thus, it is of great interest to develop a method to explore the properties of Transformers for both better local and global feature exploration while reducing the computational costs for high-quality, high-resolution image reconstruction.
%---

% %---
% In Object detection tasks, \cite{EdgeViTs} and \cite{EdgeNeXt} both combine convolutional layer and transformer layer, achieving high performance with minimal parameter and computation costs, with transformer layers combine feature in long-range pixel dependencies whereas convolutional layers concentrate on local receptive fields.
% %
% The \cite{EdgeViTs} uses a depth-wise convolutional layer as its Local Aggregation Block.
% %
% Even so, the depth-wise convolutional layer only has static weight for the entire feature and no weight-sharing, in contrast to the dynamic weight in the window attention method.
% %
% In order to solve this problem, the \cite{iDynamicDWConv} proposes a novel operation with sparse connection, weight sharing, and dynamic weight generated by a HyperNet.

%

In this paper, we propose an effective lightweight dynamic local and global self-attention network (DLGSANet) to solve SISR efficiently.
To alleviate the problem caused by the discontinuous windows, we first develop a simple yet effective multi-head dynamic local self-attention (MHDLSA) module.
The MHDLSA is motivated by the network designs of Transformers and can dynamically explore the local self-attention based on a fully CNN model to better extract local features.
As not all the tokens from the queries are relevant to those in keys, using all similarities does not effectively facilitate the high-resolution image reconstruction.
To overcome this problem, we develop a sparse global self-attention (SparseGSA) module to select the most useful similarity values for feature aggregation.
We propose a hybrid dynamic-Transformer block (HDTB) that integrates the MHDLSA and SparseGSA to explore both local and global features for high-resolution image reconstruction.
We further develop a residual hybrid dynamic-Transformer group (RHDTG) that stacks the HDTB based on the residual learning.
%%%%%
We formulate the RHDTGs into an end-to-end trainable network, named DLGSANet, to solve SISR. Figure \ref{complexity scatter} shows that the proposed DLGSANet model achieves comparable performance with fewer network parameters and lower computational costs.

The main contributions of this work are summarized as follows:
%---

%\begin{itemize}
\begin{compactitem}
	\item We propose a lightweight SISR model, called DLGSANet, to solve the SISR problem efficiently and effectively. Our analysis shows that the proposed model has fewer network parameters ($<5$M) and needs lower computational costs while generating competitive performance.
	\item We propose a simple yet effective multi-head dynamic local self-attention (MHDLSA) module to extract local features dynamically.
    \item We develop an effective sparse global self-attention module (SparseGSA) to generate better self-attention for global feature exploration.
\end{compactitem}
%\end{itemize}

%---

% =====================================================================================
%\vspace{-2mm}
\section{Related Work}
\label{sec: Related Work}

%---
\vspace{-1mm}
\noindent{\bf Conventional CNNs for SR.}
SRCNN~\cite{SRCNN} firstly introduces an effective end-to-end trainable CNN to solve the image super-resolution (SR) task.
Then, VDSR~\cite{VDSR} further improves the performance of CNNs by deepening the network and introducing residual learning, which leads to the emergence of a growing number of CNNs~\cite{FSRCNN, DRCN, DRRN, SRGAN} for SR tasks.
EDSR~\cite{EDSR} further improves PSNR results significantly by removing the unnecessary BatchNormal~\cite{batchnormal} layers.
Additionally, RCAN~\cite{RCAN} uses a channel attention mechanism to enable the network's capability of efficient feature aggregation, allowing the network to perform better with a deeper network.
Then, an increasing number of models, including SAN~\cite{SAN}, NLSA~\cite{NLSA}, and HAN~\cite{HAN}, propose a variety of attention mechanisms along spatial or channel dimensions.
% Then, a increasing number of models, such as SAN~\cite{SAN}, NLSA~\cite{NLSA}, HAN~\cite{HAN} , construct a variety of attention mechanisms along spatial or channel dimension for efficient feature aggregation ability.
%
Although these models produce significant results, a large number of parameters are required to build the network for better feature aggregation.
%---

%---
\noindent{\bf Efficient SR.}
% %
% There are several methods~\cite{CARN, IMDN, latticenet, LAPAR, ShuffleMixer} for efficient SR that are proposed in order to reduce the computation and parameter expenses.
%
Instead of aggregating on a single picture of fixed resolution, FSRCNN~\cite{FSRCNN} uses a post-upsampling approach to reduce FLOPs expenses.
To increase efficiency, CARN~\cite{CARN} applies group convolution and a cascade method to a residual network.
% %
While IMDN~\cite{IMDN} further reduces the parameters with information multi-distillation blocks.
%
%To increase efficiency, IMDN~\cite{IMDN} applies  information multi-distillation blocks to a residual network.
%
LatticeNet~\cite{IMDN} further improves the PSNR results with lattice blocks and with comparable parameter numbers and low FLOPs expenses.
Although these models are lightweight and efficient, the quality of the restored high-resolution images is not good compared to the large SR models.
%
%---

%---
\noindent{\bf Transformer-based methods for SR.}
%---
% Vision-Transformer~\cite{ViTs} introduces transformer from natural language processing to the field of image vision with great success.
%
%Compared to CNNs, the Transformer provides long range connections through stages and dynamic correlation between pixels.
% %
% But the impressive performance of the transformer comes at a significant cost.
% %
% In order to solve the issue, the \cite{SwinT} proposes a local window attention mechanism and achieve high performance at a reasonable cost.
%
Transformer-based methods~\cite{IPT, SwinIR} are proposed to solve image restoration tasks such as SR tasks.
%
%Since IPT~\cite{IPT} suffers from large parameter amount and FLOPs costs,
% SwinIR~\cite{SwinIR} perfectly translates Swin-Transformer~\cite{SwinT} to solve the low level image vision task and achieve high performance in SR task.
%
%SwinIR shows its great efficiency not only in PSNR results but also in Parameter amount and FLOPs costs compare to conventional CNN with attention mechanism.
%
SwinIR~\cite{SwinIR} uses the window-based attention mechanism to solve image SR and outperforms the CNN-based method in terms of accuracy and model complexity.
ELAN~\cite{ELAN} proposes a share attention technique to speed up the calculation in its group multi-head self-attention (GMSA).
%to overcome SwinIR's painfully slow inference time.
%
%However, ELAN's efficiency is constraied by the fact that it still relies on the window-attention process and requires a number of bigger windows to aggregate local information.
%Although ELAN's infere
%
On the other hand, with comparable parameter numbers and computational costs, SwinIR-light~\cite{SwinIR} surpasses state-of-the-art methods~\cite{IMDN, latticenet, CARN, LAPAR}.
%
% Although SwinIR-light has comparable parameter numbers and FLOPs expenses to conventional efficient SR models, it still has a slow inference speed.
%
ELAN-light~\cite{ELAN} further reduces the inference time.
%---

%---
Different from existing methods, we propose a lightweight DLGSANet which needs lower computational costs for better image SR.
\section{Proposed Method}
\label{sec: method}
The proposed lightweight dynamic local and global self-attention network (DLGSANet) mainly contains a shallow feature extraction module, six residual hybrid dynamic-Transformer groups (RHDTGs) for both local and global feature extraction, and a high-resolution image reconstruction module.

The shallow feature extraction uses a convolutional layer with a filter size of $3\times 3$ pixels to extract features from the input low-resolution image.
Each RHDTG takes the hybrid dynamic-Transformer block (HDTB) as the basic module. Moreover, the HDTB contains the multi-head dynamic local self-attention (MHDLSA) and the sparse global self-attention (SparseGSA).
The high-resolution image reconstruction module contains a convolutional layer with a filter size of $3\times 3$ pixels,  followed by a PixelShuffle~\cite{ESPCN} operation for upsampling.
Figure~\ref{network structure} shows the overview of the proposed DLGSANet for SISR.
In the following, we mainly explain the details of the MHDLSA, SparseGSA, and RHDTG.
%

%-------------------------------------------------------------------------
\subsection{Multi-head dynamic local self-attention}
\label{subsec: iDynamic-Transformer Block}

%\textbf{iDynamic-Transformer Block.}
%---
We note that window-based self-attention methods alleviate the huge computational costs of Transformers and achieve decent performance in SISR, as shown in~\cite{SwinIR} and~\cite{ELAN}.
However, the split windows cannot effectively extract features continuously and are unable to aggregate the information outside of the windows. Although the shifted windows are able to model the long-distance connections of the features in different windows, they lead to additional computational costs.

To overcome this problem, we propose a simple yet effective multi-head dynamic local self-attention (MHDLSA) based on the network designs of Transformers to extract local features effectively and efficiently.
The proposed MHDLSA first estimates spatial-variant filters to explore the local features dynamically. Then, we use the estimated filters as the dynamic local attention and apply them to the input features for better local feature aggregation.
Finally, similar to the Transformers that use a feed-forward network to improve feature representation, we apply a gated feed-forward network by~\cite{Restormer} to the aggregated features for better performance.

Specifically, given a feature $\mathbf{Y}_{in} \in \mathbb{R}^{H\times W\times C}$ generated by a layer norm followed by a $1\times 1$ convolution, we first develop a squeeze and excitation network (SENet)~\cite{SENet} without any normalize layer and non-linear activations as our dynamic weight generation network.
To ensure the generated dynamic weight better models the local information, we further use a depth-wise convolutional layer in the SENet as the depth-wise convolutional operation is able to model local attentions~\cite{convnext}.
The proposed dynamic weight generation is achieved by:
\begin{equation}
    \begin{aligned}
	&\mathbf{Y}       = \mathrm{DConv}_{7\times 7}(\mathrm{Conv}_{1\times 1}(\mathbf{Y}_{in})), \mathbf{Y} \in \mathbb{R}^{H \times W \times \gamma C} \\
	&\mathbf{Y}_{out} = \mathrm{Conv}_{1\times 1}(X), \mathbf{Y}_{out} \in \mathbb{R}^{H \times W \times G \times K^2} \\
    &\mathbf{W}(\mathrm{x}) = \mathcal{R}(\mathbf{Y}_{out}), \mathbf{W}(\mathrm{x}) \in \mathbb{R}^{G \times K \times K}
% 	W &= Reshape(X_{out}),
	\end{aligned}
	\label{eq: dynamic weight extraction HyperNet}
\end{equation}
where $\gamma$ denotes a squeezing factor; $\mathrm{DConv}_{7\times 7}$ denotes a depth-wise convolution with filter size of $7\times 7$ pixels; $\mathrm{Conv}_{1\times 1}$ denotes a convolution with a filter size of $1\times 1$ pixel; $\mathcal{R}$ denotes a reshaping function; $\mathrm{x}$ denotes the pixel index.
Each pixel has a correlated $K \times K$ dynamic kernel for dynamic convolution.

With the generated pixel-wise weight $\mathbf{W}$, we obtain the aggregated feature by:
\begin{equation}
	\hat{\mathbf{X}}^{l} = \mathbf{W} \circledast \mathbf{Y}_{in},\\
	\label{eq: iDynamicDWConv}
\end{equation}
where $\circledast$ denotes the Dynamic convolution~\cite{iDynamicDWConv} operation with weight-sharing mechanism for each channel. %within $G$ heads.

\begin{figure*}[t]
	\centering
	\includegraphics[width=1\textwidth]{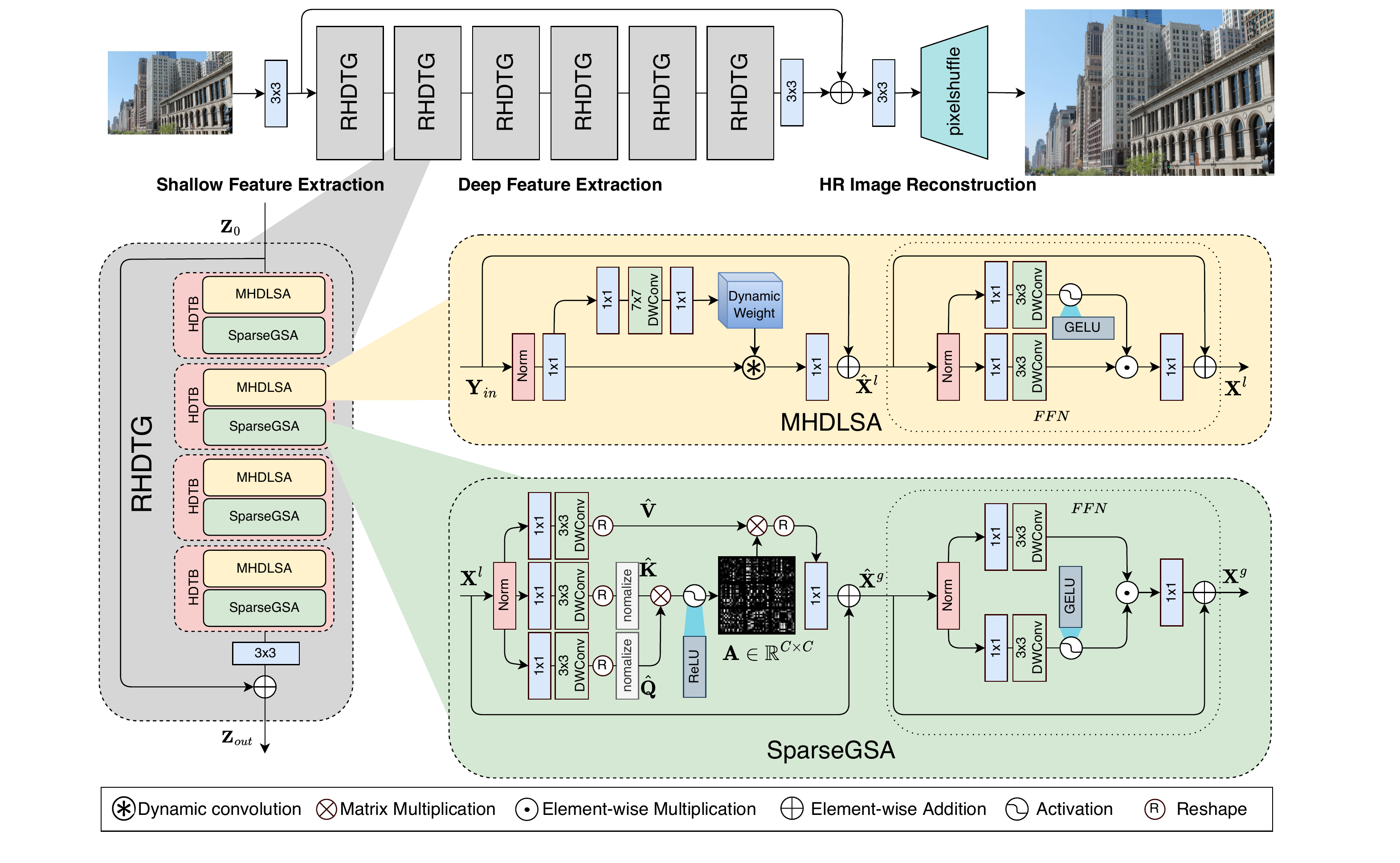} % Reduce the figure size so that it is slightly narrower than the column.
	\vspace{-10mm}
	\caption{Network architecture of the proposed DLGSANet. It mainly contains a shallow feature extraction module, six residual hybrid dynamic-Transformer groups (RHDTGs) for both local and global feature extraction, and a high-resolution image reconstruction module.
}
	\label{network structure}
    \vspace{-3mm}
\end{figure*}

The detailed network of the dynamic weight generation is shown in Figure~\ref{network structure}.
Similar to the multi-head self-attention methods~\cite{Restormer, SwinIR, SwinT}, we divide the number of feature channels into $G$ heads and learn separate dynamic weights in parallel.
%
%we set the head to $G$ and divide the channels into $\frac{C}{G}$ groups. %and learn separate dynamic weights in parallel.
%---

As the feed-forward network is widely used in Transformers for the better feature representation ability, we further apply an improved feed-forward network by~\cite{Restormer} to the aggregated feature $\hat{\mathbf{X}}$:
\begin{equation}
	\mathbf{X}^{l} = FFN(\hat{\mathbf{X}}^{l}),\\
	\label{eq: gdfn}
\end{equation}
where $FFN(\cdot)$ denotes a feed-forward network and its network details are included in Figure~\ref{network structure}.

% +++++++++++++++++++++++++++++++++++++++++++++++++++++++++++++++++++++++++++++++++++++
% +++++++++++++++++++++++++++++++++++++++++++++++++++++++++++++++++++++++++++++++++++++

%-------------------------------------------------------------------------
\subsection{Sparse global self-attention}
\label{subsec: ReLU-Transformer Block}
Although the MHDLSA is able to estimate features dynamically, it is less effective to model global features as the generated dynamic filters are based on fully convolutional operations.
Transformer-based methods are able to explore global features. However, they are usually computationally expensive.
Recent method~\cite{Restormer} develops an efficient transposed self-attention that is estimated along feature channel dimension.
Although it is efficient, the scaled dot-production attention is still generated by a softmax normalization.
We note that the softmax normalization will keep all the similarities between the tokens from the query and key.
However, not all the tokens from the queries are relevant to those in keys. Using the softmax normalization to generate self-attention would affect the following feature aggregation.
To overcome this problem, we propose a simple yet effective sparse global self-attention module.
As the ReLU is an effective activation function that can remove negative features while keeping the positive ones, we use the ReLU to keep the most useful attention for feature aggregation.
%
%By applying the ReLU to the self-attention,
%%
%With the ReLU attention map, the softmax attention map's re-weighting process transforms into activation, meaning that only a small number of channels are associated.
%%
%Moreover, ReLU activation allows for better overall network optimization during the training process.

Given a normalized feature $\mathbf{X}^{l}\in\mathbb{R}^{H \times W \times C}$ generated by the MHDLSA module, we first use a $1\times 1$ convolution followed by a $3\times 3$ depth-wise convolution to generate the query $\mathbf{Q}\in\mathbb{R}^{H \times W \times C}$, key $\mathbf{K}\in\mathbb{R}^{H \times W \times C}$, and $\mathbf{V}\in\mathbb{R}^{H \times W \times C}$.
Based on~\cite{Restormer}, we respectively apply a reshaping function to the query $\mathbf{Q}$, key $\mathbf{K}$, and value $\mathbf{V}$ and obtain $\hat{\mathbf{Q}}\in\mathbb{R}^{HW \times C}$, $\hat{\mathbf{K}}\in\mathbb{R}^{HW \times C}$, and $\hat{\mathbf{V}}\in\mathbb{R}^{HW \times C}$.
To keep the most useful attention for feature aggregation, we compute the self-attention by:
\begin{equation}
    \begin{aligned}
	\mathbf{A}= ReLU\left(\frac{\hat{\mathbf{Q}}^{\top}\hat{\mathbf{K}}}{\alpha}\right), \mathbf{A} \in \mathbb{R}^{C \times C}\\
	\end{aligned}
	\label{eq: ReLU}
\end{equation}
where $\alpha$ is a learnable parameter. Here we use the ReLU to keep the most useful attention as it is simple while can generate better results (see analysis in Section~\ref{subsec: Ablation Study}).
With the estimated attention $\mathbf{A}$, we use the same operation by~\cite{Restormer} to generate the output aggregated feature $\hat{\mathbf{X}}^g\in\mathbb{R}^{H \times W \times C}$.
Then the improved feed-forward network by~\cite{Restormer} is apply to $\hat{\mathbf{X}}^g$ to generate the output (i.e., $\mathbf{X}^g$ in Figure~\ref{network structure}).
The network details of the sparse global self-attention module are shown in Figure~\ref{network structure}.

%---
We note that using~\eqref{eq: ReLU} leads to a sparse self-attention (SparseGSA) that can keep the most useful features for high-resolution image reconstruction. The effectiveness of the proposed SparseGSA will be detailed in Section \ref{subsec: Ablation Study}.

%-------------------------------------------------------------------------
\subsection{Residual hybrid dynamic-Transformer group}
\label{subsec: Hybrid-Transformer Block}
%---
By exploring the MHDLSA and SparseGSA, we develop a hybrid dynamic-transformer block (HDTB) that contains the MHDLSA and SparseGSA for local and global feature estimations.
To reduce the training difficulty, we embed the HDTB into a residual learning framework, which leads to a hybrid dynamic-Transformer group (RHDTG).
Specifically, given the input feature $\mathbf{Z}_0$, the proposed RHDTG is achieved by:
\begin{equation}
    \begin{aligned}
	&\mathbf{Z}_{i} = \mathcal{M}_i(\mathbf{Z}_{i-1}), i=1, 2, 3, \dots, N, \\
    &\mathbf{Z}_{out} = \mathrm{Conv}_{3\times 3}(\mathbf{Z}_{N}) + \mathbf{Z}_0,
	\end{aligned}
	\label{eq: RMTG}
\end{equation}
where $\mathcal{M}_i$ denotes the $i$-th HDTB.

Finally, we formulate the proposed RHDTG into an end-to-end deep CNN model to solve SISR.
The whole network is shown in Figure~\ref{network structure}.

\begin{table*}[!t]
\centering
%\vspace{-4mm}
 \caption{Quantitative evaluations of the proposed DLGSANet against state-of-the-art methods on commonly used SISR benchmark datasets. \#Params means the number of the network parameters.
 \#FLOPs denotes the number of the FLOPs, which are calculated on images with an upscaled spatial resolution of $1280 \times 720$ pixels. Best and second best results are marked in \textcolor{red}{red} and \textcolor{blue}{blue} colors.}
 \label{tab: Classical Image Super Resolution}
\vspace{-3mm}
  \resizebox{0.95\textwidth}{!}{
  \small
   \begin{tabular}{| c | l | c | c | c | c | c | c | c |}
    \hline
    Scale & Method & \#Params(/M) & \#FLOPs(/G) & Set5 & Set14 & B100 & Urban100 & Manga109 \\
    \hline
    \multirow{9}*{$\times 2$}
    ~ & EDSR~\cite{EDSR}       & 40.73     & 9387     & 38.11/0.9602  & 33.92/0.9195  & 32.32/0.9013  & 32.93/0.9351  & 39.10/0.9773 \\
    ~ & RDN~\cite{RDN}       & 22.12     & 5098     & 38.24/0.9614  & 34.01/0.9212  & 32.34/0.9017  & 32.89/0.9353  & 39.18/0.9780 \\
    ~ & RCAN~\cite{RCAN}      & 15.44     & 3530     & 38.27/0.9614  & 34.12/0.9216  & 32.41/0.9027  & 33.34 0.9384  & 39.44/0.9786 \\
    ~ & SAN~\cite{SAN}       & 15.86     & 3050     & 38.31/0.9620  & 34.07/0.9213  & 32.42/0.9028  & 33.10/0.9370  & 39.32/0.9792 \\
    ~ & HAN~\cite{HAN}       & 63.60     & 14551    & 38.27/0.9614  & 34.16/0.9217  & 32.41/0.9027  & 33.35/0.9385  & 39.46/0.9785 \\
    ~ & NLSA~\cite{NLSA}      & 41.79     & 9632     & 38.34/0.9618  & 34.08/\textcolor{red}{0.9231}  & 32.43/0.9027  & \textcolor{blue}{33.42}/\textcolor{red}{0.9394}  & 39.59/0.9789 \\
    \cdashline{2-9}[1pt/1pt]
    ~ & SwinIR~\cite{SwinIR}    & 11.75     & 2301     & \textcolor{blue}{38.35/0.9620}  & 34.14/0.9227  & \textcolor{blue}{32.44/0.9030}  & 33.40/\textcolor{blue}{0.9393}  & \textcolor{blue}{39.60/0.9792} \\
    ~ & ELAN~\cite{ELAN}       & \textcolor{blue}{8.25}      & \textcolor{blue}{1965}     & \textcolor{red}{38.36/0.9620}  & \textcolor{blue}{34.20/0.9228}  & \textcolor{red}{32.45/0.9030}  & \textcolor{red}{33.44}/0.9391  & \textcolor{red}{39.62/0.9793} \\
    ~ & \textbf{DLGSANet (Ours)}       & \textcolor{red}{4.73}      & \textcolor{red}{1097}     & 38.34/0.9617  & \textcolor{red}{34.25/0.9231}  & 32.38/0.9025  & 33.41/\textcolor{blue}{0.9393}     & 39.57/0.9789             \\

    \hline

    \multirow{9}*{$\times 3$}
    ~ & EDSR~\cite{EDSR}       & 43.68     & 4470     & 34.65/0.9280  & 30.52/0.8462  & 29.25/0.8093  & 28.80/0.8653  & 34.17/0.9476 \\
    ~ & RDN~\cite{RDN}        & 22.30     & 2282     & 34.71/0.9296  & 30.57/0.8468  & 29.26/0.8093  & 28.80/0.8653  & 34.13/0.9484 \\
    ~ & RCAN~\cite{RCAN}      & 15.62     & 1586     & 34.74/0.9299  & 30.65/0.8482  & 29.32/0.8111  & 29.09/0.8702  & 34.44/0.9499 \\
    ~ & SAN~\cite{SAN}        & 15.89     & 1620     & 34.75/0.9300  & 30.59/0.8476  & 29.33/0.8112  & 28.93/0.8671  & 34.30/0.9494 \\
    ~ & HAN~\cite{HAN}        & 64.34     & 6534     & 34.75/0.9299  & 30.67/0.8483  & 29.32/0.8110  & 29.10/0.8705  & 34.48/0.9500 \\
    ~ & NLSA~\cite{NLSA}       & 44.74     & 4579     & 34.85/0.9306  & 30.70/0.8485  & 29.34/0.8117  & 29.25/0.8726  & 34.57 0.9508 \\
    \cdashline{2-9}[1pt/1pt]
    ~ & SwinIR~\cite{SwinIR}     & 11.93     & 1026     & 34.89/\textcolor{blue}{0.9312}  & \textcolor{blue}{30.77/0.8503}  & 29.37/\textcolor{blue}{0.8124}  & 29.29/0.8744  & \textcolor{red}{34.74}/\textcolor{red}{0.9518} \\
    ~ & ELAN~\cite{ELAN}       & \textcolor{blue}{8.27}      & \textcolor{blue}{874}      & \textcolor{blue}{34.90/0.9313}  & \textcolor{red}{30.80/0.8504}  & \textcolor{red}{29.38/0.8124}  & \textcolor{blue}{29.32/0.8745}  & 34.73/\textcolor{blue}{0.9517} \\
    ~ & \textbf{DLGSANet (Ours)}       & \textcolor{red}{4.74}      & \textcolor{red}{486}      & \textcolor{red}{34.95}/0.9310     & \textcolor{blue}{30.77}/0.8501     & \textcolor{blue}{29.38}/0.8121     & \textcolor{red}{29.43/0.8761}     & \textcolor{red}{34.76}/\textcolor{blue}{0.9517}             \\

    \hline
    \multirow{9}*{$\times 4$}
    ~ & EDSR~\cite{EDSR}       & 43.09     & 2895     & 32.46/0.8968  & 28.80/0.7876  & 27.71/0.7420  & 26.64/0.8033  & 31.02/0.9148 \\
    ~ & RDN~\cite{RDN}        & 22.27     & 1310     & 32.47/0.8990  & 28.81/0.7871  & 27.72/0.7419  & 26.61/0.8028  & 31.00/0.9151 \\
    ~ & RCAN~\cite{RCAN}       & 15.59     & 918      & 32.63/0.9002  & 28.87/0.7889  & 27.77/0.7436  & 26.82/0.8087  & 31.22/0.9173 \\
    ~ & SAN~\cite{SAN}        & 15.86     & 937      & 32.64/0.9003  & 28.92/0.7888  & 27.78/0.7436  & 26.79/0.8068  & 31.18/0.9169 \\
    ~ & HAN~\cite{HAN}        & 64.19     & 3776     & 32.64/0.9002  & 28.90/0.7890  & 27.80/0.7442  & 26.85/0.8094  & 31.42/0.9177 \\
    ~ & NLSA~\cite{NLSA}       & 44.15     & 2956     & 32.59/0.9000  & 28.87/0.7891  & 27.78/0.7444  & 26.96/0.8109  & 31.27/0.9184 \\
    \cdashline{2-9}[1pt/1pt]
    ~ & SwinIR~\cite{SwinIR}     & 11.90     & 584      & 32.72/\textcolor{blue}{0.9021}  & 28.94/\textcolor{blue}{0.7914}  & \textcolor{blue}{27.83/0.7459}  & 27.07/0.8164  & \textcolor{blue}{31.67}/\textcolor{red}{0.9226} \\
    ~ & ELAN~\cite{ELAN}       & \textcolor{blue}{8.31}      & \textcolor{blue}{494}      & \textcolor{blue}{32.75/0.9022}  & \textcolor{red}{28.96/0.7914}  & \textcolor{blue}{27.83/0.7459}  & \textcolor{blue}{27.13/0.8167}  & \textcolor{red}{31.68/0.9226} \\
    ~ & \textbf{DLGSANet (Ours)}       & \textcolor{red}{4.76}      & \textcolor{red}{274}      & \textcolor{red}{32.80}/\textcolor{blue}{0.9021}  & \textcolor{blue}{28.95}/0.7907  & \textcolor{red}{27.85/0.7464}  & \textcolor{red}{27.17/0.8175}  & \textcolor{red}{31.68}/\textcolor{blue}{0.9219}             \\

    \hline
  \end{tabular}}
\vspace{-2mm}
\end{table*}
% +++++++++++++++++++++++++++++++++++++++++++++++++++++++++++++++++++++++++++++++++++++

% =====================================================================================
\section{Experimental Results}
\label{sec: Experiments}
In this section, we perform both quantitative and qualitative evaluations to demonstrate the effectiveness of the proposed DLGSANet on commonly used benchmark datasets.
%
%-------------------------------------------------------------------------

\subsection{Experimental settings}

%--
\noindent{\bf Datasets.} We adopt the commonly used DIV2K dataset as the training dataset and evaluate our method on the commonly used test datasets, including Set5~\cite{Set5}, Set14~\cite{Set14}, B100~\cite{B100}, Urban100~\cite{Urban100}, and Manga109 ~\cite{Manga109}.
%---

%---
\noindent{\bf Implementation details.}
In the proposed DLGSANet, we use 6 RHDTGs, where each RHDTG contains 4 HDTBs. The feature channel number is set to be 90, and the multi-head number is set to be 6.
We also evaluate the proposed DLGSANet in lightweight settings by reducing the numbers of the RHDTG, the HDTB, and the feature channel.
When the numbers of the RHDTG, the HDTB, and the feature channel are set to be 3, 3, and 48, respectively, we refer to the DLGSANet as DLGSANet-tinny.
When the numbers of the RHDTG, the HDTB, and the feature channel are set to be 4, 3, and 48, respectively, we refer to the DLGSANet as DLGSANet-light.
%
%Given that the performance of the Swin-Transformer Layer is correlated with the number of channels and that SparseGSA computes attention map in the channel dimension, we can employ fewer channels with a few performance hit in order to account for parameter and calculation costs.
%
%When the channel number is set to 90, a significant parameter and calculation cost are reduced.
%
%With the aforementioned settings, our model only required 4.7M parameters while SwinIR~\cite{SwinIR} required 11M.
%
%For tiny models, the channel number, the RHDTG number, HDTB number decrease to [48, 4, 3] for Light model, [48, 3, 3] for Tiny model.
%
During the training, the mini-batch size is set to be 16. The patch size is set to be $48 \times 48$ pixels.
The initial learning rate is set to be  $5 \times 10^{-4}$ with a multi-step scheduler in 500K iterations.
We train our model using the Adam optimizer~\cite{adam} with default parameter settings.
All the networks are trained and performed using the PyTorch framework on a machine with two NVIDIA GeForce RTX 3090 GPUs.
As pointed out by~\cite{TLC}, the global attention in image restoration usually has a gap between the training and testing stages, we thus use the test-time local converter (TLC) approach by~\cite{TLC} during the testing stage.

Following the protocols used in existing methods (e.g.,~\cite{SwinIR,RCAN,EDSR}), we calculate the PSNR and SSIM scores using the Y channel in the YCbCr color space as quantitative comparisons.
Moreover, the FLOPs of each evaluated method are obtained based on upscaled images with a spatial resolution of $1280 \times 720$ pixels.

%---

% +++++++++++++++++++++++++++++++++++++++++++++++++++++++++++++++++++++++++++++++++++++
\begin{figure*}[!t]\footnotesize
 \begin{center}
  \begin{tabular}{cccccccc}
   \multicolumn{3}{c}{\multirow{5}*[51pt]{\includegraphics[width=0.38\linewidth]{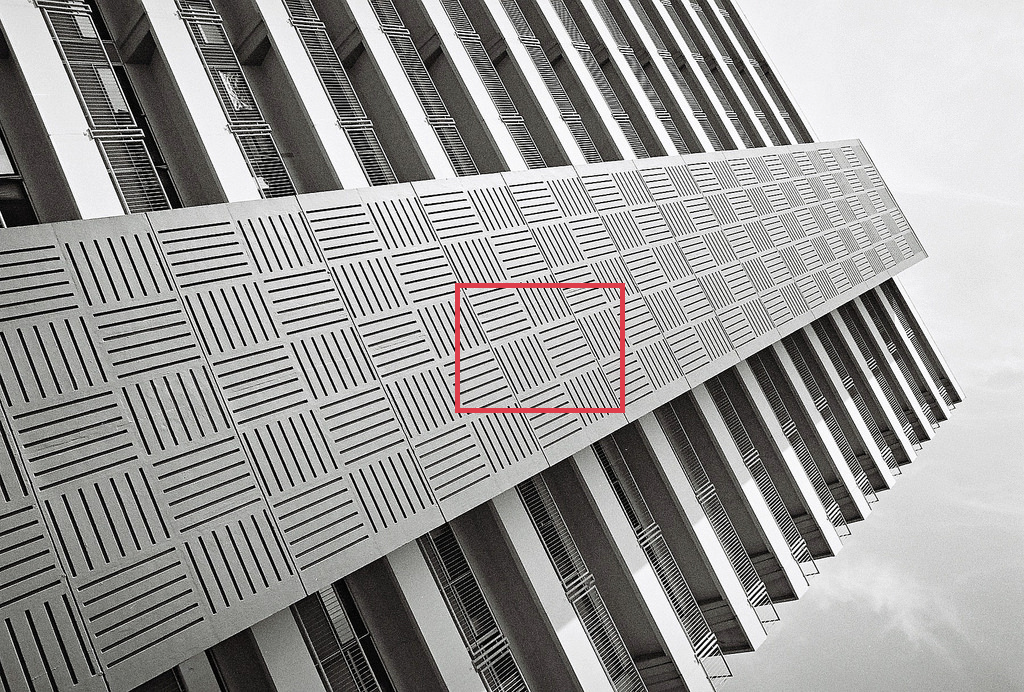}}}&\hspace{-3.5mm}
   \includegraphics[width=0.15\linewidth]{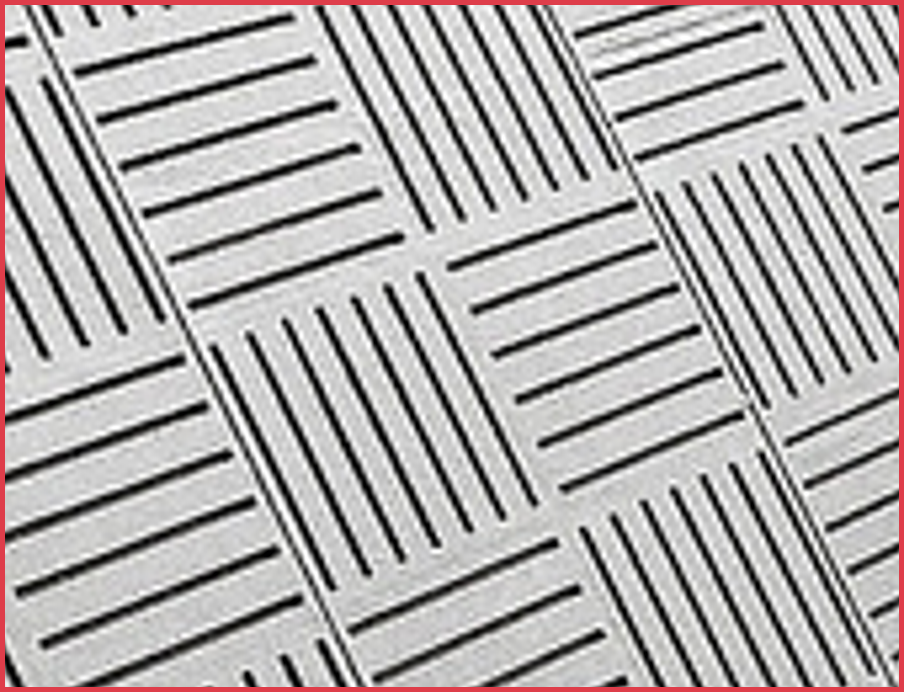} &\hspace{-3.5mm}
   \includegraphics[width=0.15\linewidth]{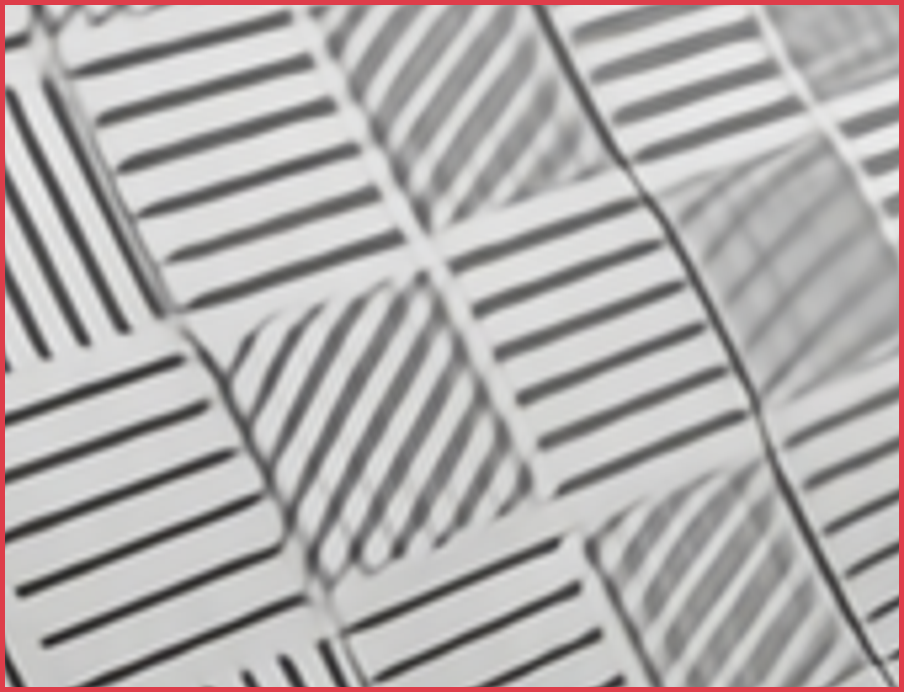} &\hspace{-3.5mm}
   \includegraphics[width=0.15\linewidth]{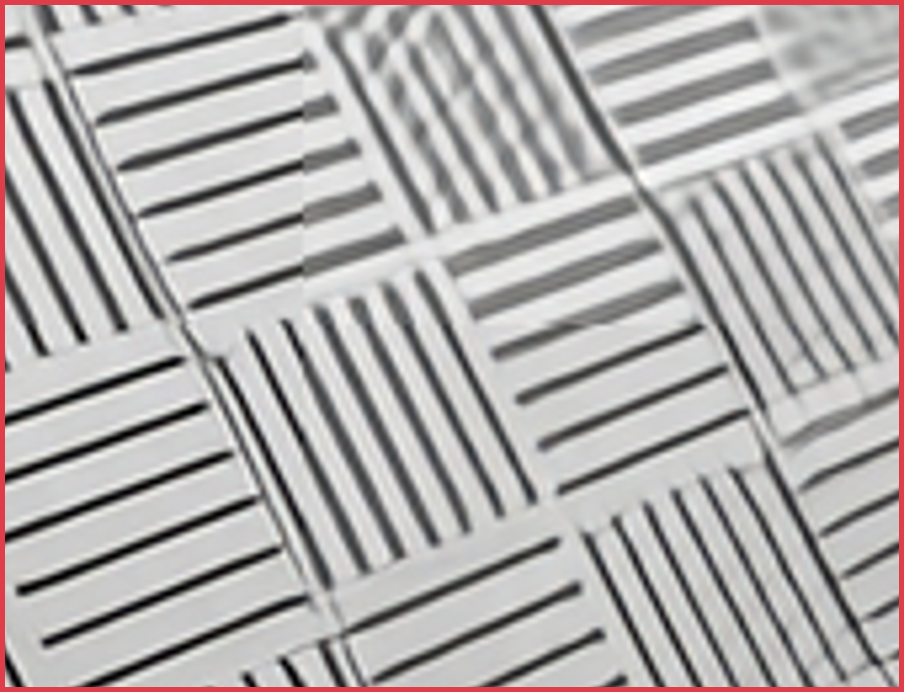} &\hspace{-3.5mm}
   \includegraphics[width=0.15\linewidth]{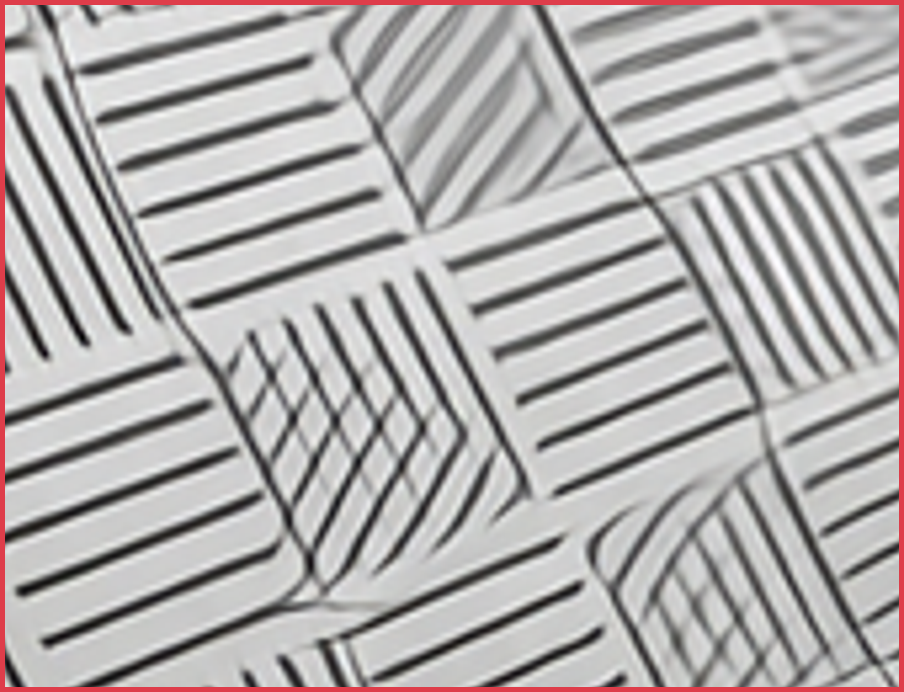} \\
   \multicolumn{3}{c}{~} &\hspace{-3.5mm}  (a) HR &\hspace{-3.5mm}  (b) EDSR~\cite{EDSR} &\hspace{-3.5mm}  (c) RCAN~\cite{RCAN}  &\hspace{-3.5mm}  (d) SAN~\cite{SAN}\\

   \multicolumn{3}{c}{~} & \hspace{-3.5mm}
   \includegraphics[width=0.15\linewidth]{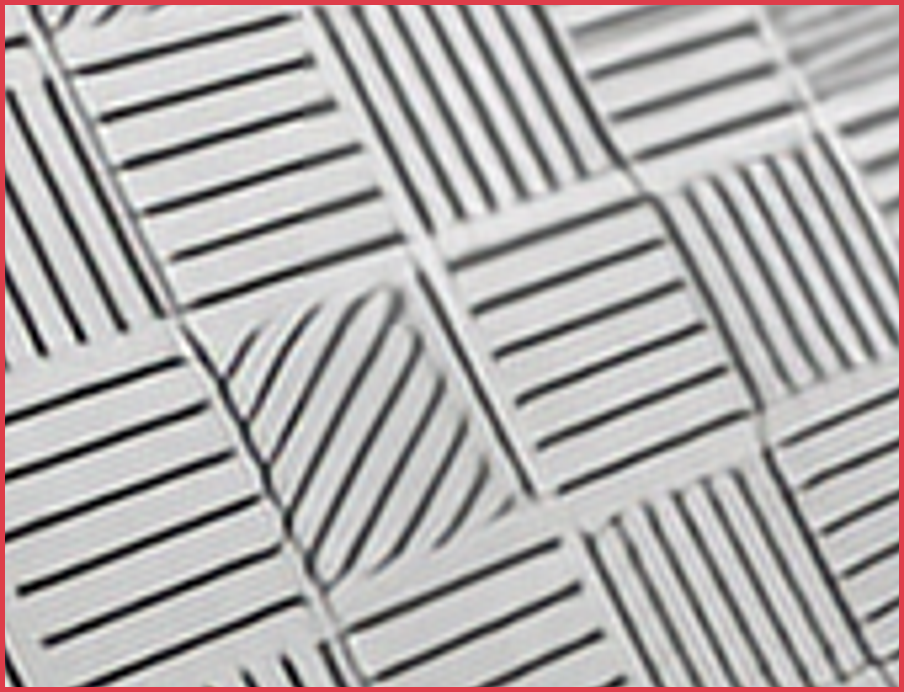} & \hspace{-3.5mm}
   \includegraphics[width=0.15\linewidth]{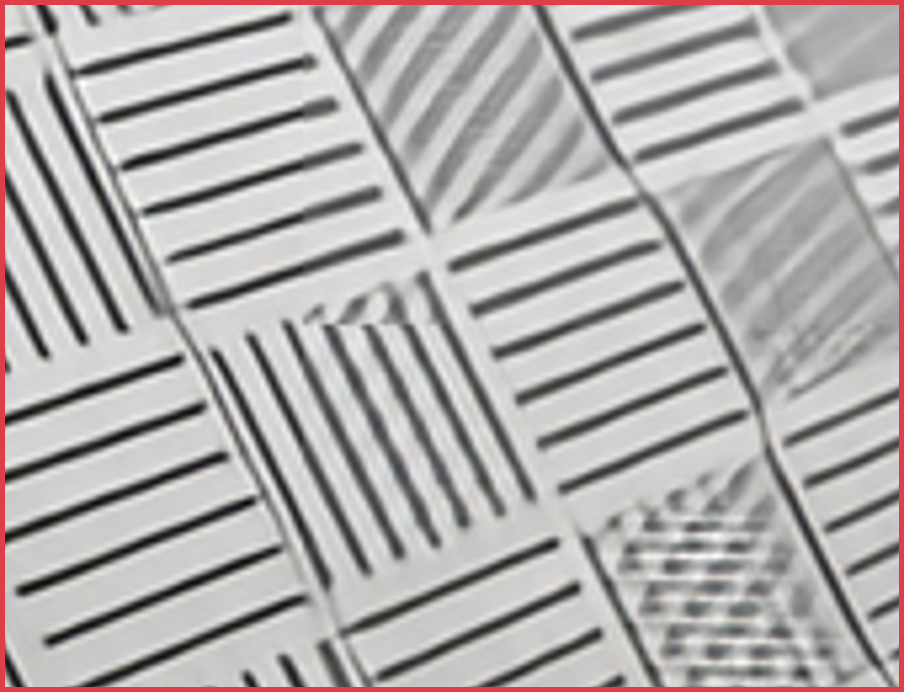} & \hspace{-3.5mm}
   \includegraphics[width=0.15\linewidth]{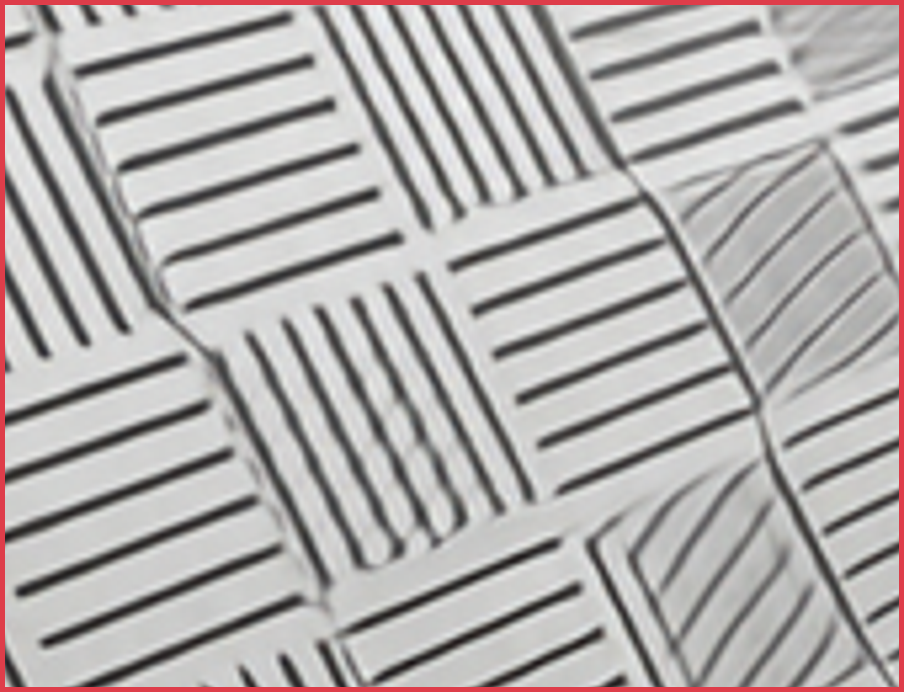} & \hspace{-3.5mm}
   \includegraphics[width=0.15\linewidth]{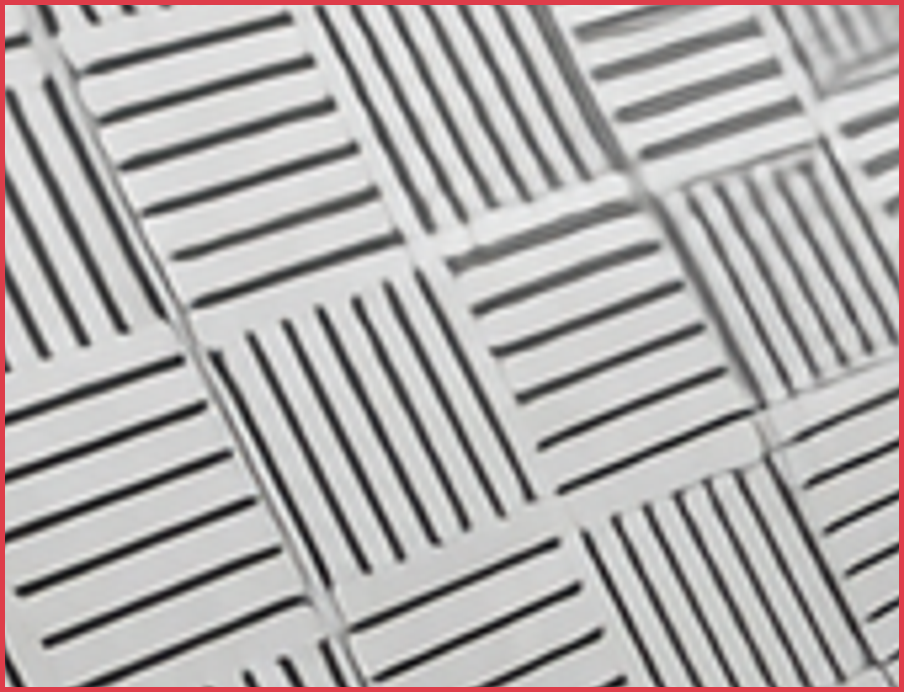} \\
   \multicolumn{3}{c}{\hspace{-3.5mm} Urban-img092 } &\hspace{-3.5mm}  (e) HAN~\cite{HAN} &  \hspace{-3.5mm} (f) NLSA~\cite{NLSA}  &\hspace{-3.5mm}  (g) SwinIR~\cite{SwinIR}  & \hspace{-3.5mm}(h) Ours \\

  \end{tabular}
 \end{center}
 \vspace{-5mm}
 %\caption{$\times 4$ results on img092 on Urban100 dataset. In the area with lines, the SR results in (b) through (g) all yield incorrect results. Due to global information extraction, our DLGSANet produces a correct image with more structural details.}
 \caption{Super-resolution results ($\times 4$) on the ``img092" image from the Urban100 dataset. The structures of the stripes are not recovered well by the evaluated methods. }
 \label{fig: results_Urban-img092}
 \vspace{-4mm}
\end{figure*}
% +++++++++++++++++++++++++++++++++++++++++++++++++++++++++++++++++++++++++++++++++++++

% +++++++++++++++++++++++++++++++++++++++++++++++++++++++++++++++++++++++++++++++++++++
\begin{figure}[!t]\footnotesize
 \begin{center}
  \begin{tabular}{cccccccc}
   \multicolumn{3}{c}{\multirow{5}*[29pt]{\includegraphics[width=0.51\linewidth]{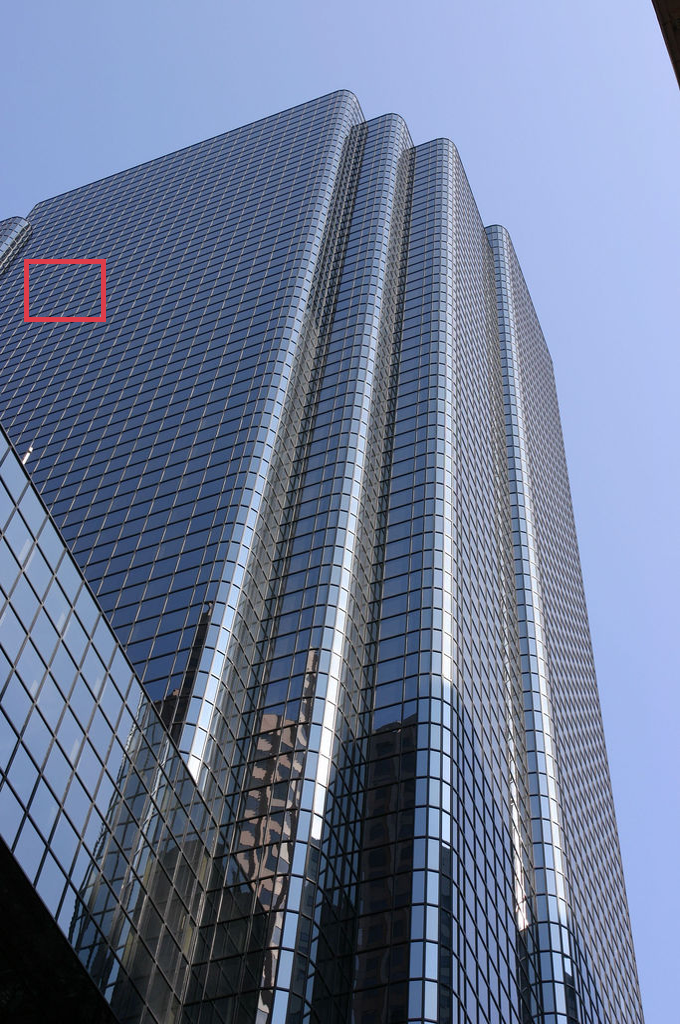}}}&\hspace{-3.5mm}
   \includegraphics[width=0.20\linewidth]{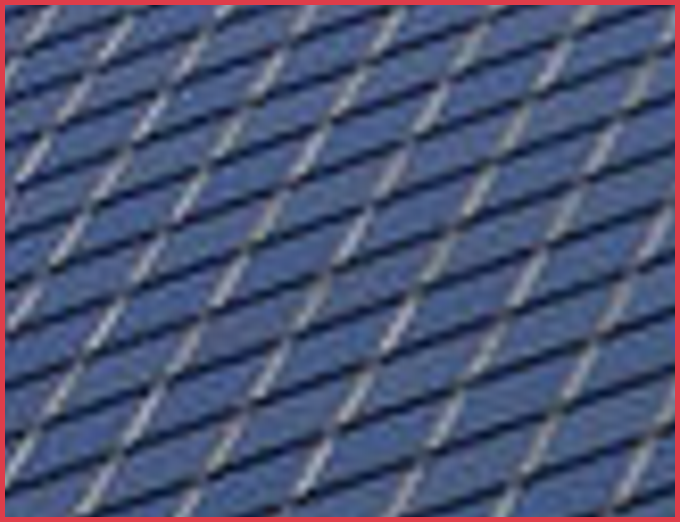} &\hspace{-3.5mm}
   \includegraphics[width=0.20\linewidth]{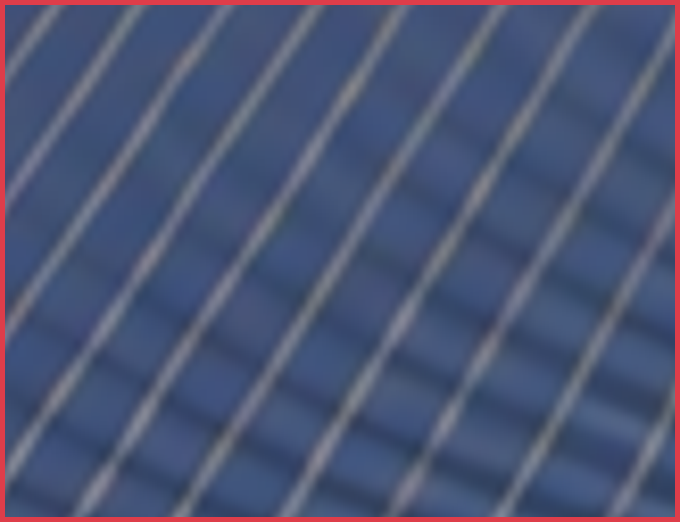} &\hspace{-3.5mm}\\

   \multicolumn{3}{c}{~} &\hspace{-3.5mm}  (a) HR               &\hspace{-3.5mm}  (b) EDSR~\cite{EDSR}  &\hspace{-3.5mm}\\

   \multicolumn{3}{c}{~} &\hspace{-3.5mm}
   \includegraphics[width=0.20\linewidth]{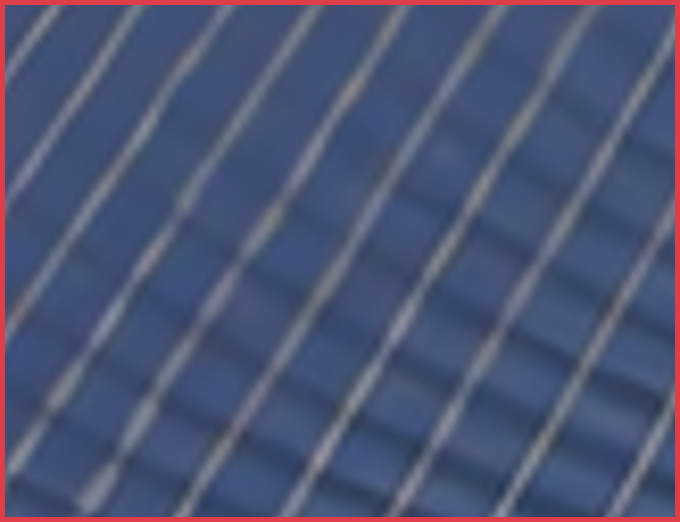} &\hspace{-3.5mm}
   \includegraphics[width=0.20\linewidth]{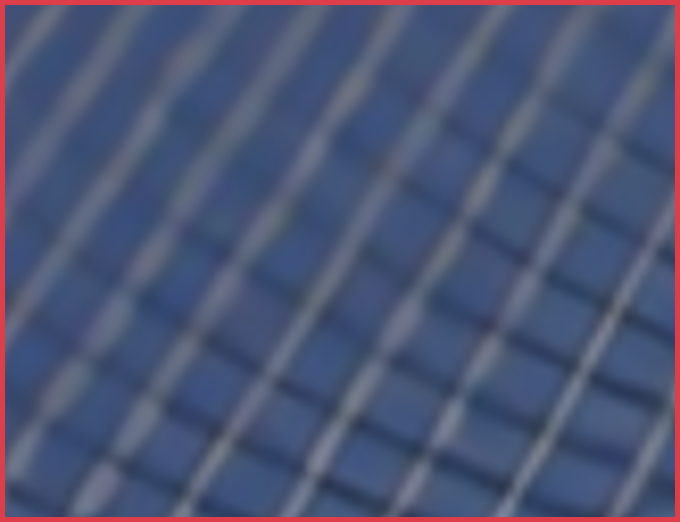}&\hspace{-3.5mm} \\

   \multicolumn{3}{c}{~} &\hspace{-3.5mm} (c) RCAN~\cite{RCAN}  &\hspace{-3.5mm}  (d) SAN~\cite{SAN}    &\hspace{-3.5mm}\\

   \multicolumn{3}{c}{~} & \hspace{-3.5mm}
   \includegraphics[width=0.20\linewidth]{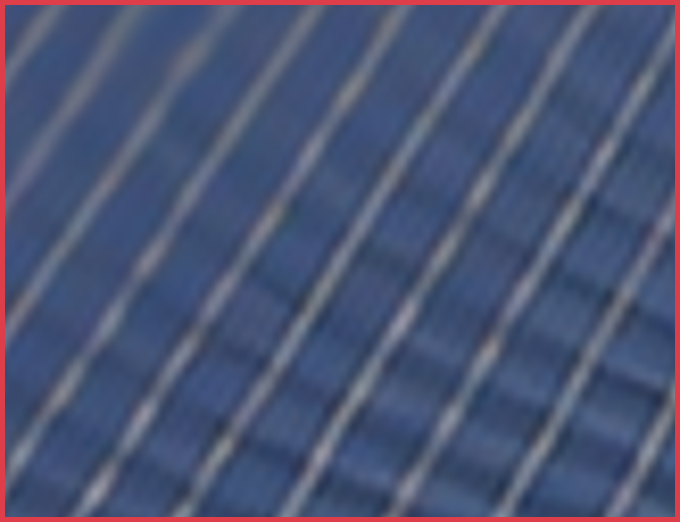} & \hspace{-3.5mm}
   \includegraphics[width=0.20\linewidth]{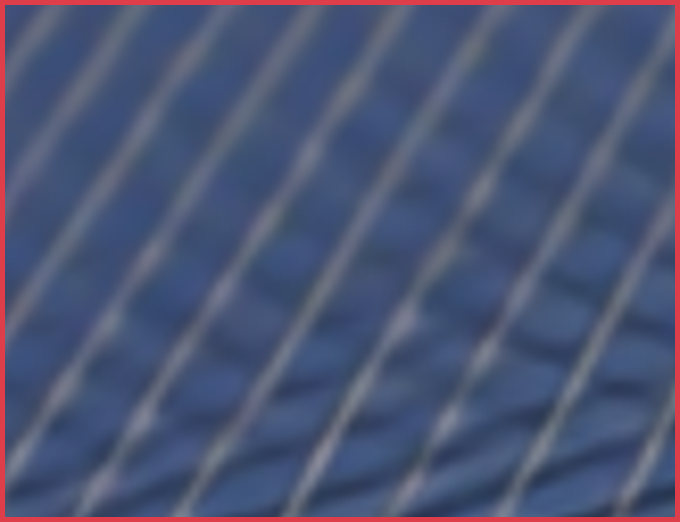} & \hspace{-3.5mm}\\

   \multicolumn{3}{c}{~} &\hspace{-3.5mm} (e) HAN~\cite{HAN}    &  \hspace{-3.5mm} (f) NLSA~\cite{NLSA}  &\hspace{-3.5mm} \\

   \multicolumn{3}{c}{~} &\hspace{-3.5mm}
   \includegraphics[width=0.20\linewidth]{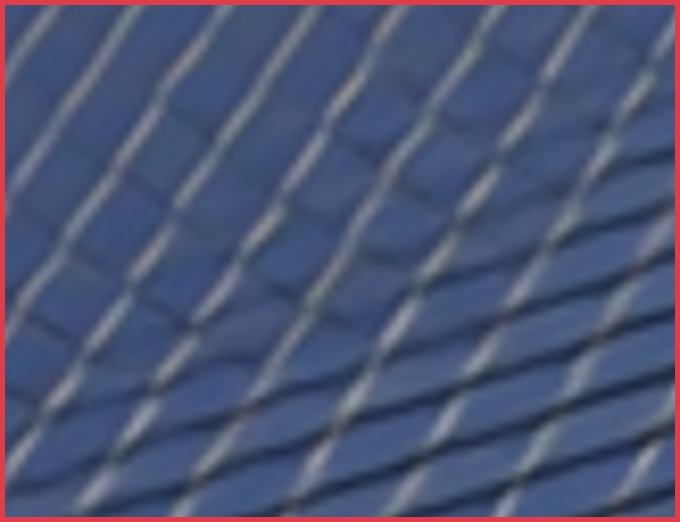} & \hspace{-3.5mm}
   \includegraphics[width=0.20\linewidth]{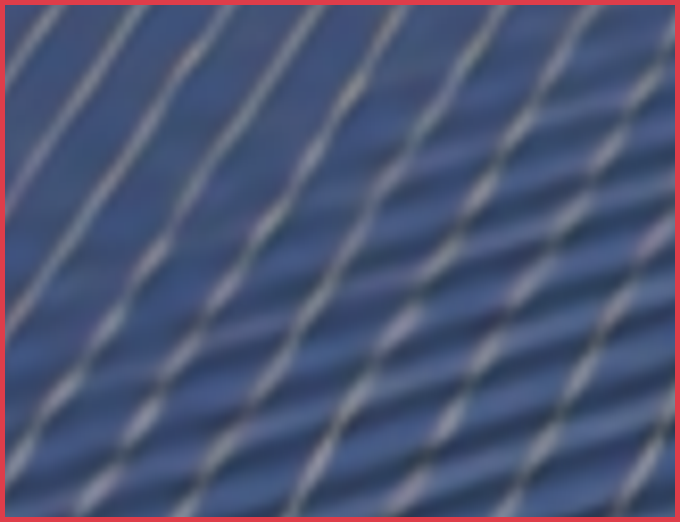} \\

   \multicolumn{3}{c}{\hspace{-3.5mm} Urban-img074 } &\hspace{-3.5mm}    (g) SwinIR~\cite{SwinIR}  & \hspace{-3.5mm}    (h) Ours &\hspace{-3.5mm}\\

  \end{tabular}
 \end{center}
 \vspace{-5mm}
 \caption{Super-resolution results ($\times 4$) on the ``img074" image from the Urban100 dataset. The evaluated methods do not recover the windows of the building well, as shown in (b)-(g).}
 \vspace{-5mm}
 \label{fig: results_Urban-img074}
\end{figure}
% +++++++++++++++++++++++++++++++++++++++++++++++++++++++++++++++++++++++++++++++++++++

%-------------------------------------------------------------------------
\subsection{Comparison results}
\label{subsec: Comparison Results}
%---
%
We compare the proposed DLGSANet with state-of-the-art methods, including SwinIR~\cite{SwinIR}, ELAN~\cite{ELAN}, NLSA~\cite{NLSA}, HAN~\cite{HAN}, RCAN~\cite{RCAN}, and EDSR~\cite{EDSR}.
%---
%---
\noindent\textbf{Quantitative evaluations.}
Table~\ref{tab: Classical Image Super Resolution} shows the quantitative evaluation results on the commonly used SR image benchmarks.
%---
We note that the proposed DLGSANet performs favorably against state-of-the-art methods in terms of network parameters and FLOPs while generating competitive results.
Particularly, compared to conventional CNN-based models, e.g., EDSR~\cite{EDSR}, the proposed DLGSANet achieves 0.62dB gains on the Urban100 dataset in terms of PSNR, while the network parameters and FLOPs of the EDSR method are $\times 10$ times than those of our DLGSANet.
Compared to the channel attention-based method~\cite{RCAN,NLSA}, our DLGSANet achieves 0.35dB and 0.21dB gains on the Urban100 dataset in terms of PSNR while utilizing $\times 3$ times and $\times 10$ times fewer parameters and FLOPs. 
%
% Our DLGSANet may focus more on local information aggregation in attention due of the dynamic weight for each pixel given by the MHDLSA, whereas typical convolutional layers treat each pixel with the same static weight.
% %
% Therefore, as seen in Figure ~\ref{fig: results_Urban-img095} in the area with white blocks, all CNN-based approaches create fuzzy results due to the static convolutional weight on local area, whereas ours can accurately reconstruct the clear individual blocks.
%
When compared to the Transformer-based methods, our DLGSANet slightly outperforms the most recent approaches, SwinIR and ELAN.
As shown in Table~\ref{tab: Classical Image Super Resolution}, DLGSANet performs better on the Urban100 when the scale factor is $\times 4$ while our method has fewer network parameters and lower FLOPs than the SwinIR method~\cite{SwinIR}.
We note that the ELAN method~\cite{ELAN} outperforms the SwinIR method~\cite{SwinIR}. However, our method still generates comparable results. More importantly, our method has fewer network parameters and lower FLOPs than the ELAN method~\cite{ELAN}.
All comparisons presented in Table~\ref{tab: Classical Image Super Resolution} show that DLGSANet is lightweight and much more efficient than the state-of-the-art methods.
% %---
%---

% +++++++++++++++++++++++++++++++++++++++++++++++++++++++++++++++++++++++++++++++++++++
\begin{table*}[!t]
 \renewcommand\arraystretch{1.}
 \begin{center}
 \caption{Quantitative evaluations of the lightweight DLGSANet against state-of-the-art methods on commonly used benchmark datasets. Best and second best results are marked in \textcolor{red}{red} and \textcolor{blue}{blue} colors. \#Params means the number of the network parameters.
 \#FLOPs denotes the number of the FLOPs which are calculated on images with an upscaled spatial resolution of $1280 \times 720$ pixels.}
 \label{tab: Small Image Super Resolution}
 \vspace{-3mm}
  \resizebox{0.95\textwidth}{!}{
  \small
   \begin{tabular}{| c | l | c | c | c | c | c | c | c |}
    \hline
    Scale & Method & \#Params(/K) & \#FLOPs(/G) & Set5 & Set14 & B100 & Urban100 & Manga109 \\
    \hline
    \multirow{6}*{$\times 2$}
      & EDSR-baseline~\cite{EDSR}  & 1370      & 316.3    & 37.99/0.9604 & 33.57/0.9175 & 32.16/0.8994 & 31.98/0.9272 & 38.54/0.9769 \\
      & IMDN~\cite{IMDN}   & 694        & 158.8     & 38.00/0.9605  & 33.63/0.9177   & 32.19/0.8996  & 32.17/0.9283   & 38.88/0.9774 \\
      & LatticeNet~\cite{latticenet}     & 756       & 169.5    & 38.06/0.9607 & 33.70/0.9187 & 32.20/0.8999 & 32.25/0.9288 & \-/\-          \\
      \cdashline{2-9}[1pt/1pt]
      & SwinIR-light~\cite{SwinIR}   & 878       & 195.6    & 38.14/\textcolor{blue}{0.9611} & 33.86/\textcolor{blue}{0.9206} & \textcolor{red}{32.31/0.9012} & 32.76/0.9340 & 39.12/\textcolor{red}{0.9783} \\
      & ELAN-light~\cite{ELAN}         & \textcolor{blue}{582}       & \textcolor{blue}{168.4}    & \textcolor{blue}{38.17/0.9611} & \textcolor{red}{33.94/0.9207} & \textcolor{blue}{32.30}/\textcolor{red}{0.9012} & 32.76/0.9340 & 39.11/\textcolor{blue}{0.9782} \\
      & \textbf{DLGSANet-tiny (Ours)}    & \textcolor{red}{566}       & \textcolor{red}{128.1}    & 38.16/\textcolor{blue}{0.9611} & \textcolor{blue}{33.92}/0.9202 & 32.26/0.9007 & \textcolor{blue}{32.82/0.9343} & \textcolor{blue}{39.14}/0.9777 \\
      & \textbf{DLGSANet-light (Ours)}  & 745       & 170      & \textcolor{red}{38.20/0.9612} & 33.89/0.9203 & \textcolor{blue}{32.30}/\textcolor{red}{0.9012} & \textcolor{red}{32.94/0.9355} & \textcolor{red}{39.29}/0.9780 \\

    \hline

    \multirow{6}*{$\times 3$}
      & EDSR-baseline~\cite{EDSR}  & 1555      & 160.2    & 34.37/0.9270 & 30.28/0.8417 & 29.09/0.8052 & 28.15/0.8527 & 33.45/0.9439 \\
      & IMDN~\cite{IMDN}  & 703  & 71.5   & 34.36/0.9270  & 30.32/0.8417  & 29.09/0.8046 & 28.17/0.8519 & 33.61/0.9445 \\
      & LatticeNet~\cite{latticenet}      & 765       & 76.3     & 34.40/0.9272 & 30.32/0.8416 & 29.10/0.8049 & 28.19/0.8513 & \-/\-          \\
      \cdashline{2-9}[1pt/1pt]
      & SwinIR-light~\cite{SwinIR}   & 886       & 87.2     & 34.62/\textcolor{blue}{0.9289} & 30.54/\textcolor{blue}{0.8463} & 29.20/0.8082 & 28.66/0.8624 & 33.98/0.9478 \\
      & ELAN-light~\cite{ELAN}         & \textcolor{blue}{590}       & 75.7     & 34.61/0.9288 & 30.55/\textcolor{blue}{0.8463} & \textcolor{blue}{29.21}/0.8081 & \textcolor{blue}{28.69}/0.8624 & 34.00/0.9478 \\
      & \textbf{DLGSANet-tiny (Ours)}    & \textcolor{red}{572}       & \textcolor{red}{56.8}     & \textcolor{blue}{34.63}/0.9288 & \textcolor{blue}{30.57}/0.8459 & \textcolor{blue}{29.21/0.8083} & \textcolor{blue}{28.69/0.8630} & \textcolor{blue}{34.10/0.9480} \\
      & \textbf{DLGSANet-light (Ours)}   & 752       & \textcolor{blue}{75.4}     & \textcolor{red}{34.70/0.9295} & \textcolor{red}{30.58/0.8465} & \textcolor{red}{29.24/0.8089} & \textcolor{red}{28.83/0.8653} & \textcolor{red}{34.16/0.9483} \\

    \hline

    \multirow{6}*{$\times 4$}
      & EDSR-baseline~\cite{EDSR}  & 1518      & 114.0    & 32.09/0.8938 & 28.58/0.7813 & 27.57/0.7357 & 26.04/0.7849 & 30.35/0.9067 \\
      & IMDN~\cite{IMDN} & 715  & 40.9 & 32.21/0.8948 & 28.58/0.7811 & 27.56/0.7353 & 26.04/0.7838 & 30.45/0.9075 \\
      & LatticeNet~\cite{latticenet}     & 777       & 43.6     & 32.18/0.8943 & 28.61/0.7812 & 27.57/0.7355 & 26.14/0.7844 & \-/\-          \\
      \cdashline{2-9}[1pt/1pt]
      & SwinIR-light~\cite{SwinIR}   & 897       & 49.6     & 32.44/0.8976 & 28.77/0.7858 & 27.69/0.7406 & 26.47/0.7980 & 30.92/\textcolor{blue}{0.9151} \\
      & ELAN-light~\cite{ELAN}         & \textcolor{blue}{601}       & 43.2     & 32.43/0.8975 & 28.78/0.7858 & 27.69/0.7406 & 26.54/0.7982 & 30.92/0.9150 \\
      & \textbf{DLGSANet-tiny (Ours)}    & \textcolor{red}{581}       & \textcolor{red}{32.0}     & \textcolor{blue}{32.46/0.8984} & \textcolor{blue}{28.79/0.7861} & \textcolor{blue}{27.70/0.7408} & \textcolor{blue}{26.55/0.8002} & \textcolor{blue}{30.98}/0.9137 \\
      & \textbf{DLGSANet-light (Ours)}   & 761       & \textcolor{blue}{42.5}     & \textcolor{red}{32.54/0.8993} & \textcolor{red}{28.84/0.7871} & \textcolor{red}{27.73/0.7415} & \textcolor{red}{26.66/0.8033} & \textcolor{red}{31.13/0.9161} \\

    \hline

  \end{tabular}}
 \end{center}
 \vspace{-6mm}
\end{table*}
% +++++++++++++++++++++++++++++++++++++++++++++++++++++++++++++++++++++++++++++++++++++

%
\noindent{\bf Qualitative evaluations.}
We compare the visual results of $\times 4$ super-resolution on the Urban100 dataset between the proposed method and state-of-the-art ones (EDSR~\cite{EDSR}, RCAN~\cite{RCAN}, SAN~\cite{SAN}, HAN~\cite{HAN}, NLSA~\cite{NLSA}, SwinIR~\cite{SwinIR}).
Figure~\ref{fig: results_Urban-img092} shows visual comparisons of the evaluted methods.
As the typical convolutional layers do not model the locally variant structures, the CNN-based methods do not correct boundaries.
The window-based self-attention methods do not effectively aggregate information outside of the windows, which thus affects the quality of the restored image (see Figure~\ref{fig: results_Urban-img092}(g)).
%
%In contrast, the proposed DLGSANet involves a MHDLSA that is able to model locally variant structural details, which thus restores better results, as shown in Figures~\ref{fig: results_Urban-img092}(h).
%
In contrast, our DLGSANet explores both local and global information by the MHDLSA and SparseGSA and restores a better image with clear blocks and boundaries, as shown in Figure~\ref{fig: results_Urban-img092}(h).

Figure~\ref{fig: results_Urban-img074} shows another visual comparison, where our method generates a better super-resolved image than the evaluated methods.

% %---
% Recent Transformer approaches SwinIR and ELAN both employ window-attention on local receptive fields and perform well on SISR.
% %
% Since window-attention only concentrates on local information and has the issue of inefficiency of discontinuous windows, DLGSANet can better gather continuous local feature and fuse global information in HDTB.
%
% Figure ~\ref{fig: results_Urban-img092} and Figure ~\ref{fig: results_Urban-img074} show that our method clearly reconstructs the proper results by aggregating the context in larger receptive fields while being more efficient in collecting global information.
%
% Window-Attenton methods, which only concentrate on small regions and can not aggregate information outside the window, are unable to rebuild as many areas as our model.
%
% Additionally, window-attention has the issue with large channels (180 in channel numbers) for better performance.
%
% Our model DLGSANet, in contrast, uses only 90 channels and can create a deeper network for greater feature aggregation.
%
% As shown in Table ~\ref{tab: Classical Image Super Resolution}, DLGSANet performs better on $\times 4$ on Urban100 (0.1 dB improved) and is about two times less in parameter number and FLOPs than SwinIR.
% %
% Even though that ELAN outperforms SwinIR with excellent efficiency, we continue to reduce the parameter and FLOPs cost, and with an improved 0.05 dB on $\times 4$ on Urban100.
% %---

\noindent{\bf Comparisons with lightweight models.}
%----
We also compare DLGSANet-tiny and DLGSANet-light with the state-of-the-art lightweight SISR models, including EDSR-baseline~\cite{EDSR}, IMDN~\cite{IMDN}, LatticNet~\cite{latticenet}, SwinIR-light~\cite{SwinIR}, and ELAN-light~\cite{ELAN}.
Table~\ref{tab: Small Image Super Resolution} shows that our proposed DLGSANet-tiny and DLGSANet-light perform better than the lightweight state-of-the-art deep models on five datasets.
Particularly, the DLGSANet-tiny has the fewest network parameters and the lowest FLOPs.
In addition, it is worth mentioning that our DLGSANet-light performs better than ELAN-light (0.21dB gains on $\times 4$ Manga109) while the DLGSANet-light has similar FLOPs to ELAN-light.

%---

%-------------------------------------------------------------------------
\section{Ablation Study and Analysis}
\label{subsec: Ablation Study}

%---
In this section, we further evaluate the effect of the components in the proposed method and compare the proposed method with baseline models.
%To ensure that all the models has the similar network parameters for fair comparisons, we train them by using the same settings in our ablation study: 90 in channel numbers, 6 in RHDTGs groups, and 4 in HDTB blocks.
For fair comparisons, we train all the baseline models using the same settings as the proposed DLGSANet.
We use the Urban100 dataset as the test dataset, as it contains a variety of images with various kinds of structural information.
%---

\begin{figure}[!t]
%\footnotesize
\scriptsize
 \begin{center}
  \begin{tabular}{cccccccc}
   \multicolumn{3}{c}{\multirow{5}*[80pt]{\includegraphics[width=0.423\linewidth]{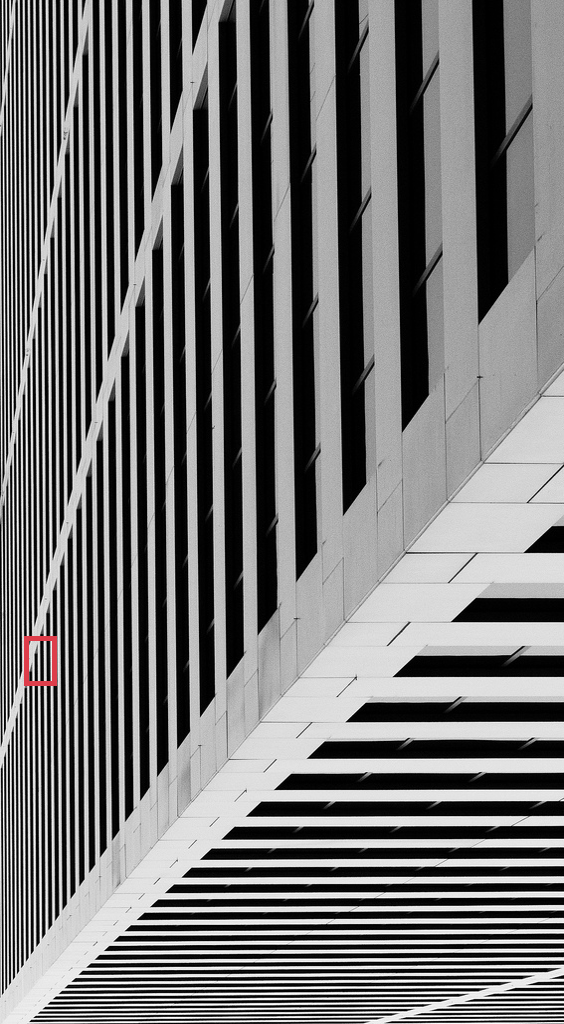}}}&\hspace{-3.5mm}
   \includegraphics[width=0.24\linewidth]{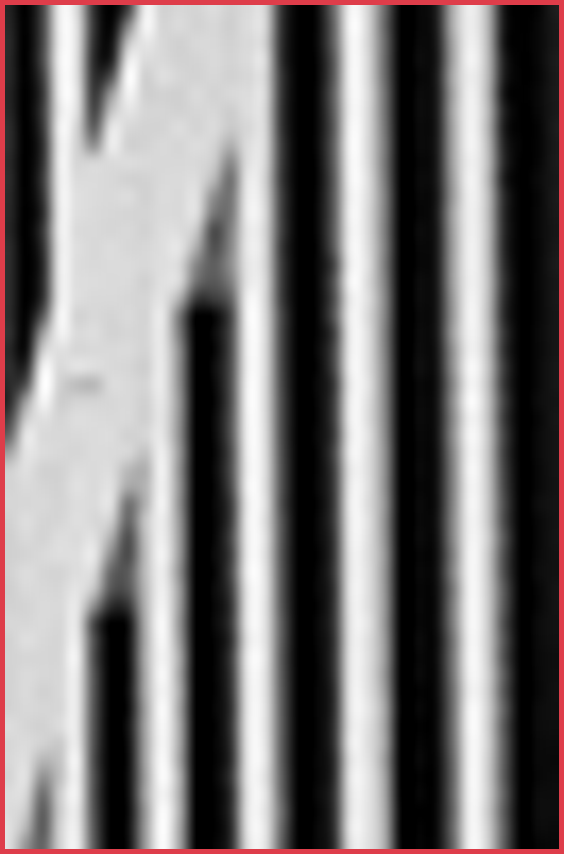} &\hspace{-3.5mm}
   \includegraphics[width=0.24\linewidth]{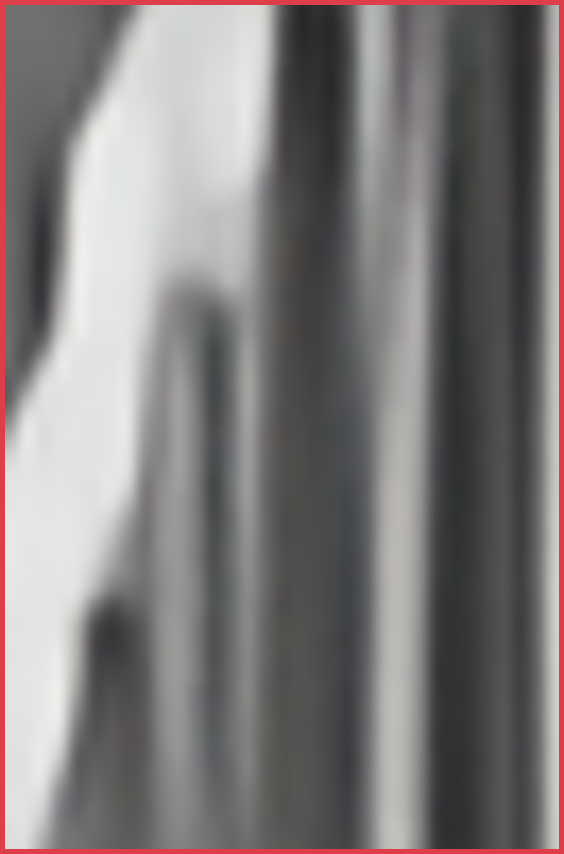} &\hspace{-3.5mm}\\
   \multicolumn{3}{c}{~} &\hspace{-3.5mm}  (a) HR &\hspace{-3.5mm}  (b) HDTB$_\textrm{SparseGSA}$ &\hspace{-3.5mm}  \\

   \multicolumn{3}{c}{~} & \hspace{-3.5mm}
   \includegraphics[width=0.24\linewidth]{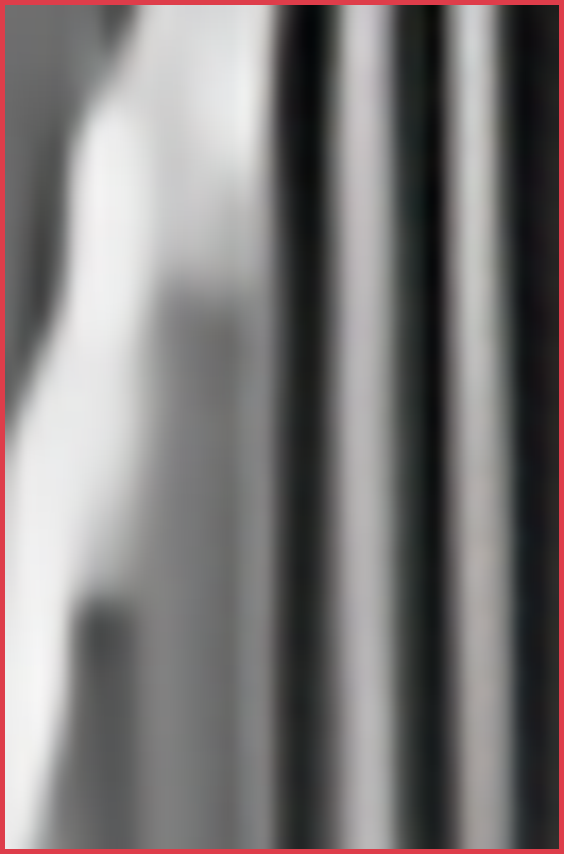} & \hspace{-3.5mm}
   \includegraphics[width=0.24\linewidth]{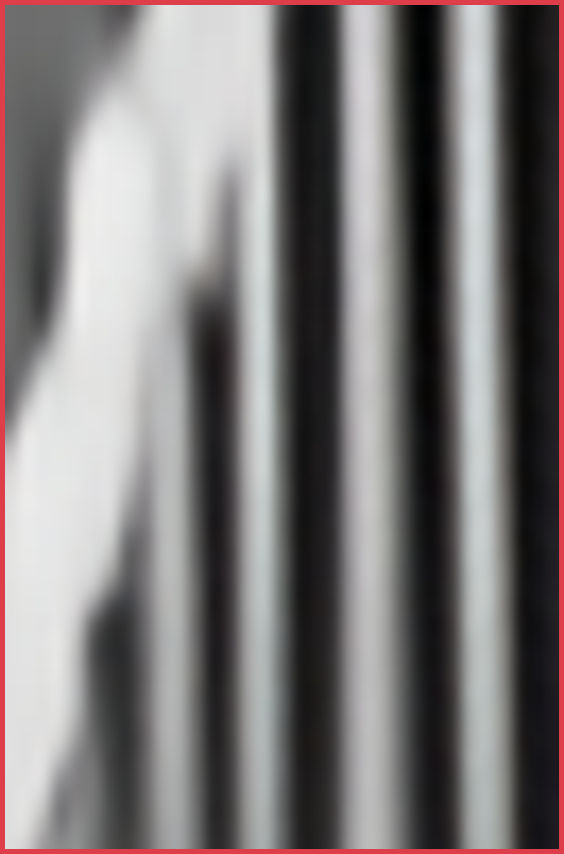} & \hspace{-3.5mm} \\
   \multicolumn{3}{c}{\hspace{-3.5mm} Urban-img011 } &\hspace{-3.5mm}  (c) HDTB$_\textrm{MHDLSA}$  &\hspace{-3.5mm}  (d) HDTB \\

  \end{tabular}
 \end{center}
 \vspace{-5mm}
 \caption{Effect of the MHDLSA and the SparseGSA in the HDTB for SISR. The results ($\times 4$) are obtained from the ``img011" image of the Urban100 dataset.}
 \label{fig: ablation_results_Urban-img011}
 \vspace{-1mm}
\end{figure}

%---
\noindent{\bf Effectiveness of the HDTB.}
As one of the key components in our DLGSANet, the HDTB fuses both local and global information for better feature aggregation.
As the HDTB contains MHDLSA and SparseGSA, we compare the proposed method with two baselines. One baseline is that we use two MHDLSA blocks in HDTB (HDTB$_\textrm{MHDLSA}$ for short). The other one is that we use two SparseGSA blocks in HDTB (HDTB$_\textrm{SparseGSA}$ for short).
The main reason we use two blocks in the HDTB is to ensure these baseline models have similar network parameters as the proposed network.
We train these two baselines using the same settings as the proposed method for fairness.
Table~\ref{tab: HDTB ablation study} shows that only using the MHDLSA generates the results with a PSNR value of 26.88dB and using the SparseGSA generates the results with a PSNR value of 26.86dB. The PSNR values of these two baselines are lower than the HDTB, suggesting the effectiveness of using both MHDLSA and SparseGSA in the HDTB for SISR.
Figure~\ref{fig: ablation_results_Urban-img011}(b) and (c) show that only using the MHDLSA or the SparseGSA in the HDTB does not restore the structures well.
In contrast, using both the MHDLSA and SparseGSA in HDTB leads to a clearer image with finer structural details (see Figure~\ref{fig: ablation_results_Urban-img011}(d)).
% Comparing to the two baselines, the model that utilize the HDTB provides clear results, as seen in the Figure~\ref{fig: ablation_results_Urban-img011}.
%
%%---------------------------------------------------------
%%\vspace{-3mm}
%\begin{table}[!t]
%  \caption{Ablation study w.r.t. the MHDLSA and SparseGSA in the HDTB. The results ($\times 4$) are obtained from the Urban100 dataset.}
%   \vspace{-3mm}
%   \label{tab: HDTB ablation study}
%\footnotesize
%%\resizebox{0.49\textwidth}{!}{
% \centering
%%  \begin{tabular}{|c|c|c|c|c|c|c|c|c|}
% \begin{tabular}{lccc}
%    \toprule
%    %\multicolumn{2}{c}{Part}                   \\
%%    \cmidrule{1-2}
%    Methods            &~~HDTB$_\textrm{MHDLSA}~~$      &~~HDTB$_\textrm{SparseGSA}$~~ &~~HDTB~~  \\
%    \hline
%    \#Param (M)         & 4.79      & 4.73     & 4.76      \\
%    PSNR                & 26.88      & 26.86     & {\bf 27.17}\\
% \bottomrule
%  \end{tabular}
%%}
%\vspace{-1mm}
%\end{table}
%%---------------------------------------------------------

\begin{table}[!t]
  \caption{Ablation study w.r.t. the MHDLSA and SparseGSA in the HDTB. The results ($\times 4$) are obtained from the Urban100 dataset.}
  \vspace{-3mm}
    \centering
  \label{tab: HDTB ablation study}
\footnotesize
\resizebox{0.48\textwidth}{!}{
 \centering
%  \begin{tabular}{|c|c|c|c|c|c|c|c|c|}
 \begin{tabular}{lcccc}
    \toprule
    %\multicolumn{2}{c}{Part}                   \\
%    \cmidrule{1-2}
  Model & MHDLSA & SparseGSA  & \#Param & PSNR\\
    \hline
        HDTB$_\textrm{MHDLSA}$ & \ding{51} &            & 4.79M   & 26.88 \\
        HDTB$_\textrm{SparseGSA}$     &           & \ding{51}  & 4.73M    & 26.86 \\
        HDTB  & \ding{51} & \ding{51}  & 4.76M   & {\bf 27.17} \\
 \bottomrule
  \end{tabular}
}
\vspace{-8mm}
\end{table}
%---------------------------------------------------------

% \begin{table}[!t]
%   \caption{Ablation study w.r.t. the MHDLSA and SparseGSA in the HDTB. The results ($\times 4$) are obtained from the Urban100 dataset.}
%   \vspace{-3mm}
%     \centering
%   \label{tab: HDTB ablation study}
% \footnotesize
% \resizebox{0.48\textwidth}{!}{
%  \centering
% %  \begin{tabular}{|c|c|c|c|c|c|c|c|c|}
%  \begin{tabular}{lcccc}
%     \toprule
%     %\multicolumn{2}{c}{Part}                   \\
% %    \cmidrule{1-2}
%   Model & MHDLSA & SparseGSA  & \#Param & PSNR\\
%     \hline
%         HDTB$_\textrm{MHDLSA}$ & \ding{51} &            & 4.79M   & 26.88 \\
%         HDTB$_\textrm{SparseGSA}$     &           & \ding{51}  & 4.73M    & 26.86 \\
%         HDTB  & \ding{51} & \ding{51}  & 4.76M   & {\bf 27.17} \\
%  \bottomrule
%   \end{tabular}
% }
% \vspace{-5mm}
% \end{table}
% %---------------------------------------------------------

% +++++++++++++++++++++++++++++++++++++++++++++++++++++++++++++++++++++++++++++++++++++
% +++++++++++++++++++++++++++++++++++++++++++++++++++++++++++++++++++++++++++++++++++++

%---
\noindent{\bf Effectiveness of the MHDLSA.}
Our MHDLSA approach inherits the property of convolution and can generate dynamic weights for better local feature exploration.
To demonstrate the effectiveness of the proposed MHDLSA, we first replace the MHDLSA with the commonly used multi-head window attention (MHSA) in the proposed network and train this baseline using the same settings as the proposed network for fair comparisons.
Table~\ref{tab: MHDLSA} shows that using the MHDLSA achieves 0.28dB gains in terms of PSNR compared to the method using the MHSA, suggesting the effectiveness of the MHDLSA on SISR.

\begin{table}[!t]
  \caption{Effectiveness of the proposed SparseGSA. The results ($\times 4$) are obtained from the Urban100 dataset.}
  \vspace{-3mm}
    \centering
  \label{tab: SparseGSA ablation study}
\footnotesize
\resizebox{0.48\textwidth}{!}{
 \centering
%  \begin{tabular}{|c|c|c|c|c|c|c|c|c|}
 \begin{tabular}{lcccc}
    \toprule
Model & ~~~~~Softmax~~~~~ & ~~~~~ReLU~~~~~ & ~~~~~\#Param~~~~~  & ~~~~~PSNR~~~~~\\
        \hline
        GSA &   \ding{51}      &       & 4.76M & 27.05 \\
        SparseGSA &   & \ding{51} & 4.76M & {\bf 27.17} \\
 \bottomrule
  \end{tabular}
}
\vspace{-4mm}
\end{table}
%---------------------------------------------------------

% +++++++++++++++++++++++++++++++++++++++++++++++++++++++++++++++++++++++++++++++++++++
\begin{figure*}[!t]
	\centering
	\includegraphics[width=.98\textwidth]{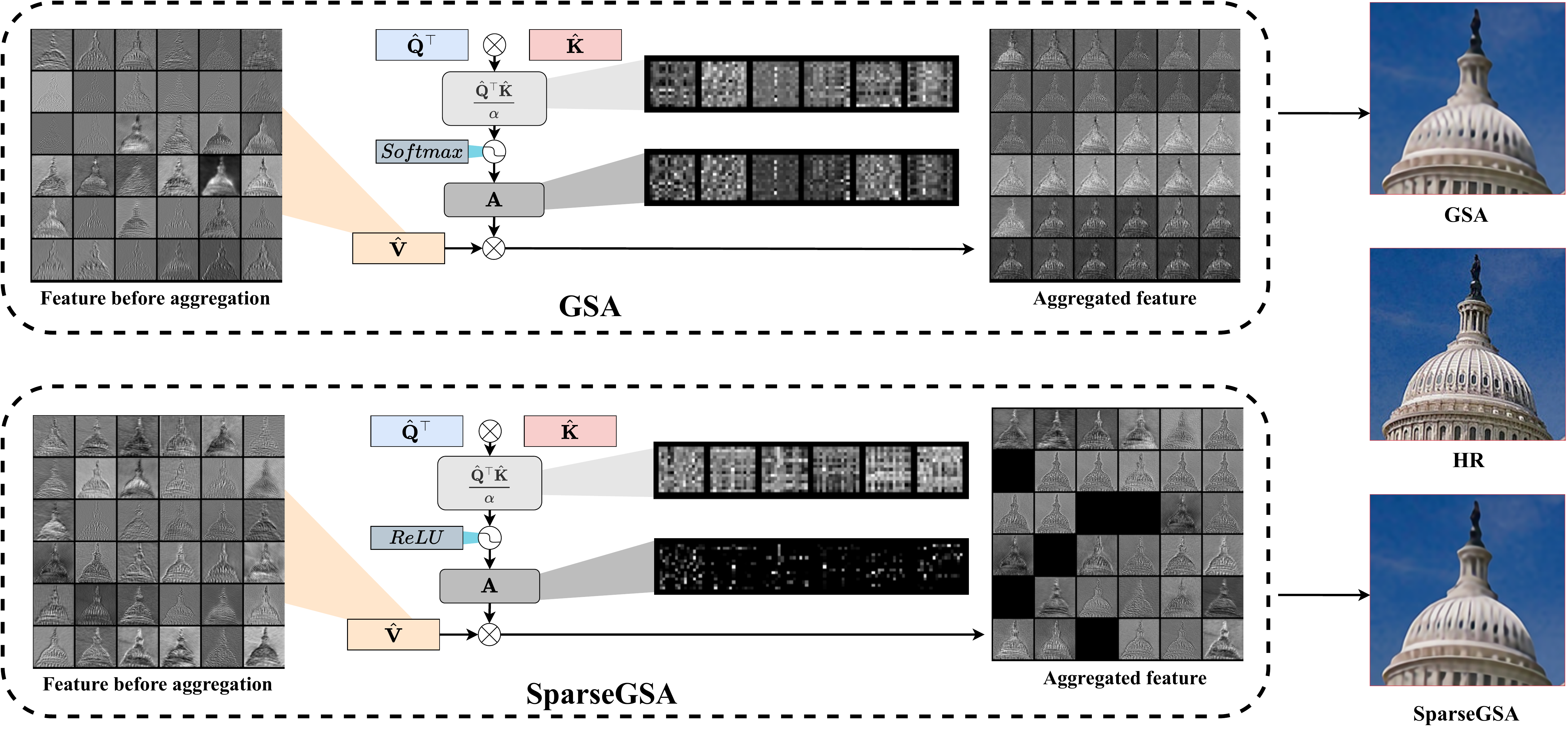} % Reduce the figure size so that it is slightly narrower than the column.
	\vspace{-3mm}
	\caption{Effect of the SparseGSA on SISR. Using the SparseGSA is able to remove useless self-attention values and thus leads to better features for high-resolution image reconstruction. }
	\label{softmaxrelu}
    \vspace{-2mm}
\end{figure*}
% +++++++++++++++++++++++++++++++++++++++++++++++++++++++++++++++++++++++++++++++++++++

\begin{table}[!t]
  \caption{Effect of the MHDLSA on SISR. The results ($\times 4$) are obtained from the Urban100 dataset.}
  \vspace{-3mm}
    \centering
  \label{tab: MHDLSA}
\footnotesize
\resizebox{0.48\textwidth}{!}{
 \centering
%  \begin{tabular}{|c|c|c|c|c|c|c|c|c|}
 \begin{tabular}{lcccccc}
    \toprule
     Model  & MHSA & MHDLSA & SparseGSA & \#Param  & PSNR\\
    \hline
        w/ MHSA     &  \ding{51}  &        &  \ding{51} & 4.67M   & 26.89 \\
        w/ MHDLSA &      &   \ding{51}    &    \ding{51}  &  4.76M   & {\bf 27.17} \\
 \bottomrule
  \end{tabular}
}
\vspace{-1mm}
\end{table}
%---------------------------------------------------+++++++++++++++++++++++++++++++++
\begin{table}[!t]
 \renewcommand\arraystretch{1.}
 \centering
  \caption{Evaluations of running time (/ms) on NVIDIA GeForce RTX 3090 GPUs. }
  \vspace{-3mm}
  \resizebox{0.46\textwidth}{!}{
  \small
   \begin{tabular}{| c | l | c | c | c |}
    \hline
    Type & Model                 & x2   & x3  & x4 \\
    \hline
    \multirow{7}{*}{Lightweight ($<1M$)}
    & EDSR-baseline~\cite{EDSR}         & 40   & 21  & 15 \\
    & IMDN~\cite{IMDN}                  & 29   & 13  & 8  \\
    & LatticNet~\cite{latticenet}             & 36   & 17  & 10 \\
    \cdashline{2-5}[1pt/1pt]
    & SwinIR-light~\cite{SwinIR}          & 340  & 145 & 81 \\ %\hline
    & ELAN-light~\cite{ELAN}            & 165  & 78  & 46 \\ %\hline
    & \textbf{DLGSANet-tiny (Ours)}  & 143  & 66  & 38 \\ %\hline
    & \textbf{DLGSANet-light (Ours)} & 192  & 88  & 51 \\ %\hline
    \hline
    \multirow{6}{*}{Regular}
    & EDSR~\cite{EDSR}                  & 679  & 344 & 232 \\
    & RCAN~\cite{RCAN}                  & 487  & 220 & 133 \\
    & NLSA~\cite{NLSA}                  & 1208 & 548 & 343 \\
    \cdashline{2-5}[1pt/1pt]
    & SwinIR~\cite{SwinIR}                & 1314 & 528 & 278 \\ %\hline
    & ELAN~\cite{ELAN}                  & 965  & 422 & 243 \\ %\hline
    & \textbf{DLGSANet (Ours)}       & 748  & 337 & 187 \\
    \hline
  \end{tabular}}
 \vspace{-3mm}
 \label{tab: Latency time}
 \vspace{-1mm}
\end{table}
% +++++++++++++++++++++++++++++++++++++++++++++++++++++++++++++++++++++++++++++++++++++

\noindent{\bf Effectiveness of the SparseGSA.}
The proposed SparseGSA uses the ReLU to remove useless self-attention for better feature aggregation.
We demonstrate the effectiveness of the SparseGSA by comparing it with the commonly used method that adopts the softmax operation.
Table~\ref{tab: SparseGSA ablation study} demonstrates that the SparseGSA outperforms the commonly used method that uses the softmax for self-attention, where the PSNR value of the method using the SparseGSA is 0.12dB higher.

We further show visualization results in Figure~\ref{softmaxrelu} to better illustrate the effect of the proposed SparseGSA.
We note that using the softmax function will keep all the self-attention values for the feature aggregation. However, if the tokens from the query and key are different, using the self-attention values of these tokens may affect the feature aggregation.
In contrast, using the ReLU removes some self-attention values. For example, only the ones that correspond to the main structures and details are preserved, which thus leads to better results, as shown in Figure~\ref{softmaxrelu}.

\noindent{\bf Running time analysis.}
We further evaluate the running time of the proposed DLGSANet against the state-of-the-art methods by using a machine with an NVIDIA GeForce RTX 3090 GPU.
We use test images with the upscaled spatial resolution of $1280 \times 720$ pixels.
Table~\ref{tab: Latency time} shows that our method, including both the regular model and the lightweight model, is more efficient than the Transformer-based methods.

%%%%%%%%%
%\vspace{-2mm}
\section{Conclusion}
%\vspace{-2mm}
   %
   We have presented an effective lightweight dynamic local and global self-attention networks (DLGSANet) to solve image super-resolution.
   The proposed DLGSANet is mainly composed of several residual hybrid dynamic-Transformer groups (RHDTGs), where each RHDTG takes the hybrid dynamic-Transformer block (HDTB) as the basic module.
   The HDTB includes a simple yet effective multi-head dynamic local self-attention module (MHDLSA) for local feature extraction and a sparse global self-attention (SparseGSA) module for global feature extraction.
   In contrast to existing Transformers, the proposed HDTB not only extracts local features efficiently but also aggregates the most useful global features by a sparse global self-attention estimation method.
   By training the proposed DLGSANet in an end-to-end manner, we show that it has fewer network parameters and lower computational costs while achieving competitive performance against state-of-the-art ones on benchmarks in terms of accuracy.

%%%%%%%%% REFERENCES
{\small
\bibliographystyle{ieee_fullname}
\bibliography{egbib}

\begin{thebibliography}{10}\itemsep=-1pt

\bibitem{CARN}
Namhyuk Ahn, Byungkon Kang, and Kyung-Ah Sohn.
\newblock Fast, accurate, and lightweight super-resolution with cascading
  residual network.
\newblock In {\em ECCV}, 2018.

\bibitem{B100}
Pablo Arbel{\'a}ez, Michael Maire, Charless~C. Fowlkes, and Jitendra Malik.
\newblock Contour detection and hierarchical image segmentation.
\newblock {\em PAMI}, 33(5):898--916, 2011.

\bibitem{Set5}
Marco Bevilacqua, Aline Roumy, Christine Guillemot, and Marie line
  Alberi~Morel.
\newblock Low-complexity single-image super-resolution based on nonnegative
  neighbor embedding.
\newblock In {\em BMVC}, 2012.

\bibitem{IPT}
Hanting Chen, Yunhe Wang, Tianyu Guo, Chang Xu, Yiping Deng, Zhenhua Liu, Siwei
  Ma, Chunjing Xu, Chao Xu, and Wen Gao.
\newblock Pre-trained image processing transformer.
\newblock In {\em CVPR}, 2021.

\bibitem{TLC}
Xiaojie Chu, Liangyu Chen, , Chengpeng Chen, and Xin Lu.
\newblock Improving image restoration by revisiting global information
  aggregation.
\newblock In {\em ECCV}, 2022.

\bibitem{SAN}
Tao Dai, Jianrui Cai, Yongbing Zhang, Shu-Tao Xia, and Lei Zhang.
\newblock Second-order attention network for single image super-resolution.
\newblock In {\em CVPR}, 2019.

\bibitem{SRCNN}
Chao Dong, Chen~Change Loy, Kaiming He, and Xiaoou Tang.
\newblock Learning a deep convolutional network for image super-resolution.
\newblock In {\em ECCV}, 2014.

\bibitem{FSRCNN}
Chao Dong, Chen~Change Loy, and Xiaoou Tang.
\newblock Accelerating the super-resolution convolutional neural network.
\newblock In {\em ECCV}, 2016.

\bibitem{ViTs}
Alexey Dosovitskiy, Lucas Beyer, Alexander Kolesnikov, Dirk Weissenborn,
  Xiaohua Zhai, Thomas Unterthiner, Mostafa Dehghani, Matthias Minderer, Georg
  Heigold, Sylvain Gelly, et~al.
\newblock An image is worth 16x16 words: Transformers for image recognition at
  scale.
\newblock In {\em ICLR}, 2020.

\bibitem{iDynamicDWConv}
Qi Han, Zejia Fan, Qi Dai, Lei Sun, Ming-Ming Cheng, Jiaying Liu, and Jingdong
  Wang.
\newblock On the connection between local attention and dynamic depth-wise
  convolution.
\newblock In {\em ICLR}, 2022.

\bibitem{OISR}
Xiangyu He, Zitao Mo, Peisong Wang, Yang Liu, Mingyuan Yang, and Jian Cheng.
\newblock Ode-inspired network design for single image super-resolution.
\newblock In {\em CVPR}, 2019.

\bibitem{SENet}
Jie Hu, Li Shen, and Gang Sun.
\newblock Squeeze-and-excitation networks.
\newblock In {\em CVPR}, 2018.

\bibitem{Urban100}
Jia-Bin Huang, Abhishek Singh, and Narendra Ahuja.
\newblock Single image super-resolution from transformed self-exemplars.
\newblock In {\em CVPR}, 2015.

\bibitem{IMDN}
Zheng Hui, Xinbo Gao, Yunchu Yang, and Xiumei Wang.
\newblock Lightweight image super-resolution with information
  multi-distillation network.
\newblock In {\em ACM MM}, 2019.

\bibitem{batchnormal}
Sergey Ioffe and Christian Szegedy.
\newblock Batch normalization: Accelerating deep network training by reducing
  internal covariate shift.
\newblock In {\em ICML}, 2015.

\bibitem{VDSR}
Jiwon Kim, Jung~Kwon Lee, and Kyoung~Mu Lee.
\newblock Accurate image super-resolution using very deep convolutional
  networks.
\newblock In {\em CVPR}, 2016.

\bibitem{DRCN}
Jiwon Kim, Jung~Kwon Lee, and Kyoung~Mu Lee.
\newblock Deeply-recursive convolutional network for image super-resolution.
\newblock In {\em CVPR}, 2016.

\bibitem{adam}
Diederik~P Kingma and Jimmy Ba.
\newblock Adam: A method for stochastic optimization.
\newblock In {\em ICLR}, 2015.

\bibitem{SRGAN}
Christian Ledig, Lucas Theis, Ferenc Husz{\'a}r, Jose Caballero, Andrew
  Cunningham, Alejandro Acosta, Andrew Aitken, Alykhan Tejani, Johannes Totz,
  Zehan Wang, et~al.
\newblock Photo-realistic single image super-resolution using a generative
  adversarial network.
\newblock In {\em CVPR}, 2017.

\bibitem{LAPAR}
Wenbo Li, Kun Zhou, Lu Qi, Nianjuan Jiang, Jiangbo Lu, and Jiaya Jia.
\newblock {LAPAR}: Linearly-assembled pixel-adaptive regression network for
  single image super-resolution and beyond.
\newblock In {\em NeurIPS}, 2020.

\bibitem{SwinIR}
Jingyun Liang, Jiezhang Cao, Guolei Sun, Kai Zhang, Luc Van~Gool, and Radu
  Timofte.
\newblock {SwinIR}: Image restoration using swin transformer.
\newblock In {\em ICCV Workshops}, 2021.

\bibitem{EDSR}
Bee Lim, Sanghyun Son, Heewon Kim, Seungjun Nah, and Kyoung~Mu Lee.
\newblock Enhanced deep residual networks for single image super-resolution.
\newblock In {\em CVPR Workshops}, 2017.

\bibitem{SwinT}
Ze Liu, Yutong Lin, Yue Cao, Han Hu, Yixuan Wei, Zheng Zhang, Stephen Lin, and
  Baining Guo.
\newblock Swin transformer: Hierarchical vision transformer using shifted
  windows.
\newblock In {\em ICCV}, 2021.

\bibitem{convnext}
Zhuang Liu, Hanzi Mao, Chao-Yuan Wu, Christoph Feichtenhofer, Trevor Darrell,
  and Saining Xie.
\newblock A convnet for the 2020s.
\newblock In {\em CVPR}, 2022.

\bibitem{latticenet}
Xiaotong Luo, Yuan Xie, Yulun Zhang, Yanyun Qu, Cuihua Li, and Yun Fu.
\newblock Latticenet: Towards lightweight image super-resolution with lattice
  block.
\newblock In {\em ECCV}, 2020.

\bibitem{Manga109}
Yusuke Matsui, Kota Ito, Yuji Aramaki, Toshihiko Yamasaki, and Kiyoharu Aizawa.
\newblock Sketch-based manga retrieval using manga109 dataset.
\newblock {\em arXiv preprint arXiv:1510.04389}, 2015.

\bibitem{NLSA}
Yiqun Mei, Yuchen Fan, and Yuqian Zhou.
\newblock Image super-resolution with non-local sparse attention.
\newblock In {\em CVPR}, 2021.

\bibitem{HAN}
Ben Niu, Weilei Wen, Wenqi Ren, Xiangde Zhang, Lianping Yang, Shuzhen Wang,
  Kaihao Zhang, Xiaochun Cao, and Haifeng Shen.
\newblock Single image super-resolution via a holistic attention network.
\newblock In {\em ECCV}, 2020.

\bibitem{ESPCN}
Wenzhe Shi, Jose Caballero, Ferenc Husz{\'a}r, Johannes Totz, Andrew~P Aitken,
  Rob Bishop, Daniel Rueckert, and Zehan Wang.
\newblock Real-time single image and video super-resolution using an efficient
  sub-pixel convolutional neural network.
\newblock In {\em CVPR}, 2016.

\bibitem{DRRN}
Ying Tai, Jian Yang, and Xiaoming Liu.
\newblock Image super-resolution via deep recursive residual network.
\newblock In {\em CVPR}, 2017.

\bibitem{Restormer}
Syed~Waqas Zamir, Aditya Arora, Salman Khan, Munawar Hayat, Fahad~Shahbaz Khan,
  and Ming-Hsuan Yang.
\newblock Restormer: Efficient transformer for high-resolution image
  restoration.
\newblock In {\em CVPR}, 2022.

\bibitem{Set14}
Roman Zeyde, Michael Elad, and Matan Protter.
\newblock On single image scale-up using sparse-representations.
\newblock In {\em Curves and Surfaces}, 2012.

\bibitem{ELAN}
Xindong Zhang, Hui Zeng, Shi Guo, and Lei Zhang.
\newblock Efficient long-range attention network for image super-resolution.
\newblock In {\em ECCV}, 2022.

\bibitem{RCAN}
Yulun Zhang, Kunpeng Li, Kai Li, Lichen Wang, Bineng Zhong, and Yun Fu.
\newblock Image super-resolution using very deep residual channel attention
  networks.
\newblock In {\em ECCV}, 2018.

\bibitem{RDN}
Yulun Zhang, Yapeng Tian, Yu Kong, Bineng Zhong, and Yun Fu.
\newblock Residual dense network for image super-resolution.
\newblock In {\em CVPR}, 2018.

\end{thebibliography}
}

\end{document}